\documentclass[]{gewu}

\usepackage{wasysym}

\usepackage[toc,page,header]{appendix}

\usepackage{minitoc}
\usepackage{microtype}      
\usepackage[dvipsnames,table]{xcolor}         

\usepackage{wrapfig}
\usepackage{hyperref}
\crefname{figure}{Fig.}{Figs.}
\usepackage{url}
\usepackage{booktabs}
\usepackage{multirow}
\usepackage{subcaption}
\usepackage{caption}
\usepackage{amssymb}
\usepackage{graphicx}
\usepackage{amsmath}
\usepackage{wrapfig}
\usepackage{enumitem}
\usepackage{arydshln} 
\usepackage{gensymb}
\usepackage{makecell}
\usepackage{pifont} 

\usepackage{tabularx}

\usepackage{booktabs}

\usepackage{array}
\usepackage{xurl} 
\usepackage{tcolorbox}
\tcbset{
  mybox/.style={
    colback  = gray!20,
    coltext  = black,
    colframe = black,
    boxrule  = 1pt,
    left     = 2mm,
    right    = 2mm,
    top      = 1mm,
    bottom   = 1mm,
    fontupper= \footnotesize, 
  }
}

\usepackage[table]{xcolor} 

\newcommand{\graycell}{\cellcolor{gray!25}}

\title{MIBench: Evaluating LMMs on Multimodal Interaction}

\author[1,2,\dagger,*]{Yu Miao}
\author[1,2,\dagger,*]{Zequn Yang}
\author[1,2]{Yake Wei}
\author[1,2]{Ziheng Chen}
\author[3,*]{Haotian Ni} 
\author[4]{\\Haodong Duan}
\author[4]{Kai Chen}
\author[1,2,\textnormal{\Letter}]{Di Hu}

\affiliation[1]{Gaoling School of Artificial Intelligence, Renmin University of China, Beijing, China}
\affiliation[2]{Beijing Key Laboratory of Research on Large Models and Intelligent Governance, Beijing, China}
\affiliation[3]{Beihang University, Beijing, China}
\affiliation[4]{Shanghai Artificial Intelligence Laboratory, Shanghai, China}

\contribution[\dagger]{Equal contribution}
\contribution[*]{Intern at Shanghai AI Lab}
\contribution[\textnormal{\Letter}]{ Corresponding author}

\abstract{
In different multimodal scenarios, it needs to integrate and utilize information across modalities in a specific way based on the demands of the task. 
Different integration ways between modalities are referred to as ``multimodal interaction”. How well a model handles various multimodal interactions largely characterizes its multimodal ability. 
In this paper, we introduce \textbf{MIBench}, a comprehensive benchmark designed to evaluate the multimodal interaction capabilities of Large Multimodal Models (LMMs), which formulates each instance as a ($con_v$, $con_t$, $task$) triplet with \textbf{con}texts from \textbf{v}ision and \textbf{t}ext, necessitating that LMMs employ correct forms of multimodal interaction to effectively complete the task.
MIBench assesses models from three key aspects: the ability to source information from vision-centric or text-centric cues, and the ability to generate new information from their joint synergy. 
Each interaction capability is evaluated hierarchically across three cognitive levels: Recognition, Understanding, and Reasoning. MIBench comprises over 10,000 vision-text context pairs spanning 32 distinct tasks.
Evaluation of state-of-the-art LMMs show that: (1) LMMs' ability on multimodal interaction remains constrained, despite the scaling of model parameters and training data; (2) they are easily distracted by textual modalities when processing vision information; (3) they mostly possess a basic capacity for multimodal synergy; and (4) natively trained multimodal models show noticeable deficits in fundamental interaction ability. 
We expect that these observations can serve as a reference for developing LMMs with more enhanced multimodal ability in the future.
}

\MyEmail{Yu Miao at \email{ymiao@ruc.edu.com}, Di Hu at \email{dihu@ruc.edu.com}}

\checkdata[Project Page]{\url{https://gewu-lab.github.io/MIBench}}

\begin{document}
\maketitle


\vspace{-0.5cm}
\section{Introduction}

Multimodal intelligence is one of the keys to fully understanding the real world. This capability requires models to integrate and utilize information from diverse sources to handle complex tasks~\cite{baltruvsaitis2018multimodal}. Crucially, the way information is presented and processed varies significantly depending on the task. Some tasks rely primarily on understanding a specific modality (e.g., Visual Question Answering based heavily on the given image), while others demand deep collaboration, where key insights emerge only from the synergy between modalities (e.g., combining images and text to interpret a chart). These different strategies for processing multimodal information are known as \textbf{multimodal interaction}~\cite{liang2023quantifying}. To achieve effective multimodal interaction\footnote{This paper focuses on the interaction between vision and text.}, three core abilities should be focused on: sourcing crucial cues from the vision modality, from the text modality, and deriving new information from the synergy between vision and text. The expected multimodal intelligence should possess all three sub-abilities simultaneously, extracting and synthesizing information comprehensively to handle an increasingly diverse range of tasks.

\begin{figure*}[!t]
    \centering
    \includegraphics[width=1\textwidth]{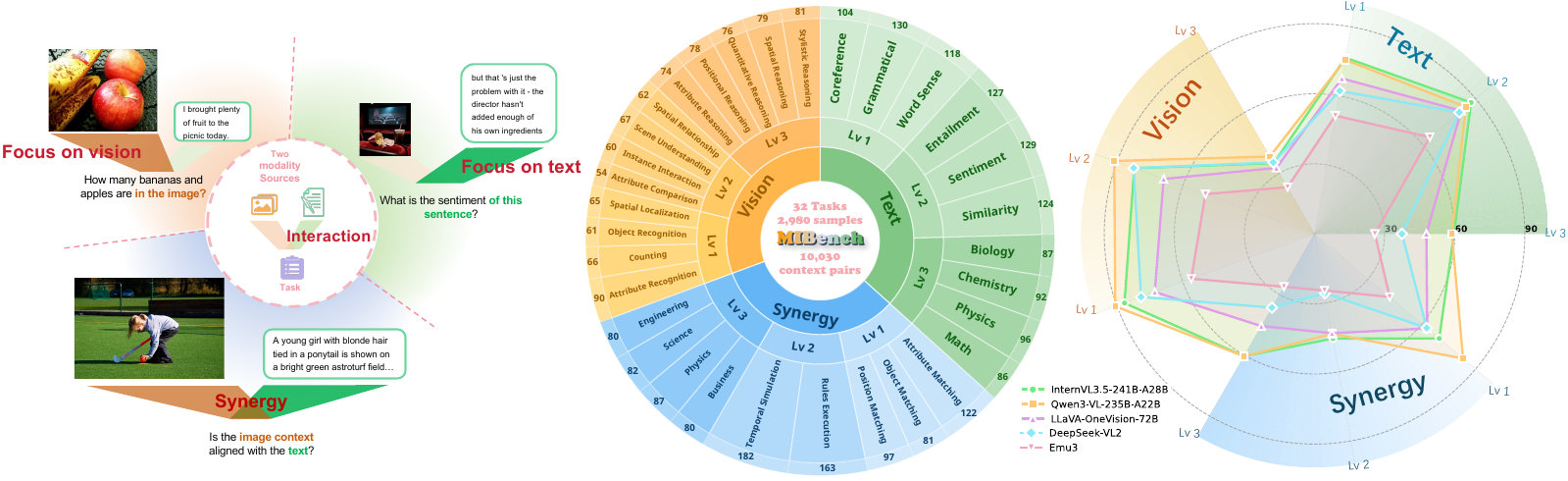}
    \vspace{-1em}
     \caption{
\textbf{(Left)} Different multimodal tasks require different types of modality interaction. Some tasks depend on information mostly from one modality (visual or textual context), while others require synergistic collaboration between them. \textbf{(Middle)} We introduce MIBench to systematically evaluate how well Large Multimodal Models (LMMs) handle multimodal interactions, which is structured around three fundamental interaction patterns (Vision-centric, Text-centric, and Synergy) across three cognitive levels (Recognition, Understanding, and Reasoning). \textbf{(Right)} Most current LMMs demonstrate regrettably limited capabilities in multimodal interaction (especially in synergy), struggling to selectively utilize cues from centric modality and achieve cross-modal collaboration.
     }
     \label{fig:teaser}
\end{figure*} 

Large Multimodal Models (LMMs) are widely regarded as a primary approach to realizing this intelligence and have recently shown impressive capabilities as general-purpose solvers~\cite{openai-gpt5.1, comanici2025gemini25pushingfrontier, Qwen2.5-VL, wang2025internvl3_5, wu2024deepseekvl2mixtureofexpertsvisionlanguagemodels}. To fully understand the strengths and limitations of these models, rigorous evaluation is essential. Early benchmarks focused mostly on visual perception~\cite{liu2024mmbench, fu2023mme, chen2024we}, often checking if models actually use visual data rather than just taking textual shortcuts~\cite{zhang2024mathverse, zheng2025mllms}. More recent studies have expanded to complex, knowledge-intensive tasks~\cite{lu2022learn, yue2024mmmu} and cross-modal reasoning~\cite{lu2024mathvista, qiao2024we, hao2025mllmsreasonmultimodalityemma}. However, these existing evaluations tend to treat multiple modalities as a single entity, neglecting the influence of interactions between information from different modality sources.
As a result, it remains unclear whether LMMs have truly mastered flexible forms of multimodal interaction. Specifically, it is uncertain whether they can selectively extract targeted information from text or images, 
and whether they can synthesize two modalities to achieve genuine cross-modal synergy. 
Consequently, critical questions remain: Do current model architectures and training paradigms enable such effective interactions? If not, where exactly do the bottlenecks lie?

To address this gap, we introduce \textbf{MIBench}, a benchmark designed to holistically assess the ability to process multimodal interaction. 
We evaluate an LMM's ability to handle three core interaction patterns for image-text tasks: sourcing visual information for \textbf{Vision-centric} tasks, sourcing textual information for \textbf{Text-centric} tasks, and achieving cross-modal \textbf{Synergy}. To enable this fine-grained analysis, we designed a new sample format: in addition to the task, each sample includes both a visual context and a textual context, which is ($con_v$, $con_t$, $task$), which allows us to verify if the model selectively leverages cues from the required source to achieve effective interaction.
For instance, to evaluate the model's ability to extract key visual information within multimodal contexts to answer questions about the image, we pair the image ($con_v$) with additional textual contexts ($con_t$) that exhibit diverse levels of relevance to the image. 
Thus, we can vary textual contexts to evaluate whether the model can truly focus on the image, or struggles with this type of interaction, making its responses sensitive to the varying textual contexts. 
And we designed a hierarchical framework that stratifies these interaction patterns across progressive cognitive levels, ranging from \textbf{Recognition} to \textbf{Understanding} and \textbf{Reasoning}. This dual categorization (by interaction pattern and cognitive level) provides a comprehensive, hierarchical structure for assessing how well models master these interaction abilities.
Based on this framework, we curated MIBench (with over 30 tasks and 10,000 visual-textual context pairs, as shown in Fig.~\ref{fig:teaser}) through rigorous collection and annotation, ensuring both the breadth and reliability of the assessment.

Using MIBench, we evaluate the interaction capabilities of current state-of-the-art LMMs, and the analysis reveals several key findings:
\vspace{-1mm}
\begin{itemize}[nosep,leftmargin=*]
   \item Current LMMs remain constrained in their ability to perform effective multimodal interactions, despite increases in model scale and data sources.
   \item LMMs demonstrate limited proficiency in selectively extracting cues from the target modality. This issue is particularly severe in vision-centric tasks, as the models often fail to prioritize visual evidence over the textual context.
   \item While LMMs show acceptable performance in basic cross-modal alignment, their effectiveness drops in tasks requiring deep interactive understanding and reasoning, for which parameter scaling is insufficient to address.
   \item  Native LMMs struggle with fundamental perception, creating a bottleneck for complex interactions. Non-native LMMs appear to lean on powerful LLM foundations for better synergy but are hindered by a strong text bias that prevents deeper cross-modal collaboration
\end{itemize}

By focusing on multimodal interaction, MIBench offers fresh insights into whether current LMMs are truly achieving multimodal intelligence. It also provides clear direction and inspiration for the next steps in developing more advanced, genuine synergistic capabilities.

\begin{figure*}[!t]
    \centering
    \includegraphics[width=1\linewidth]{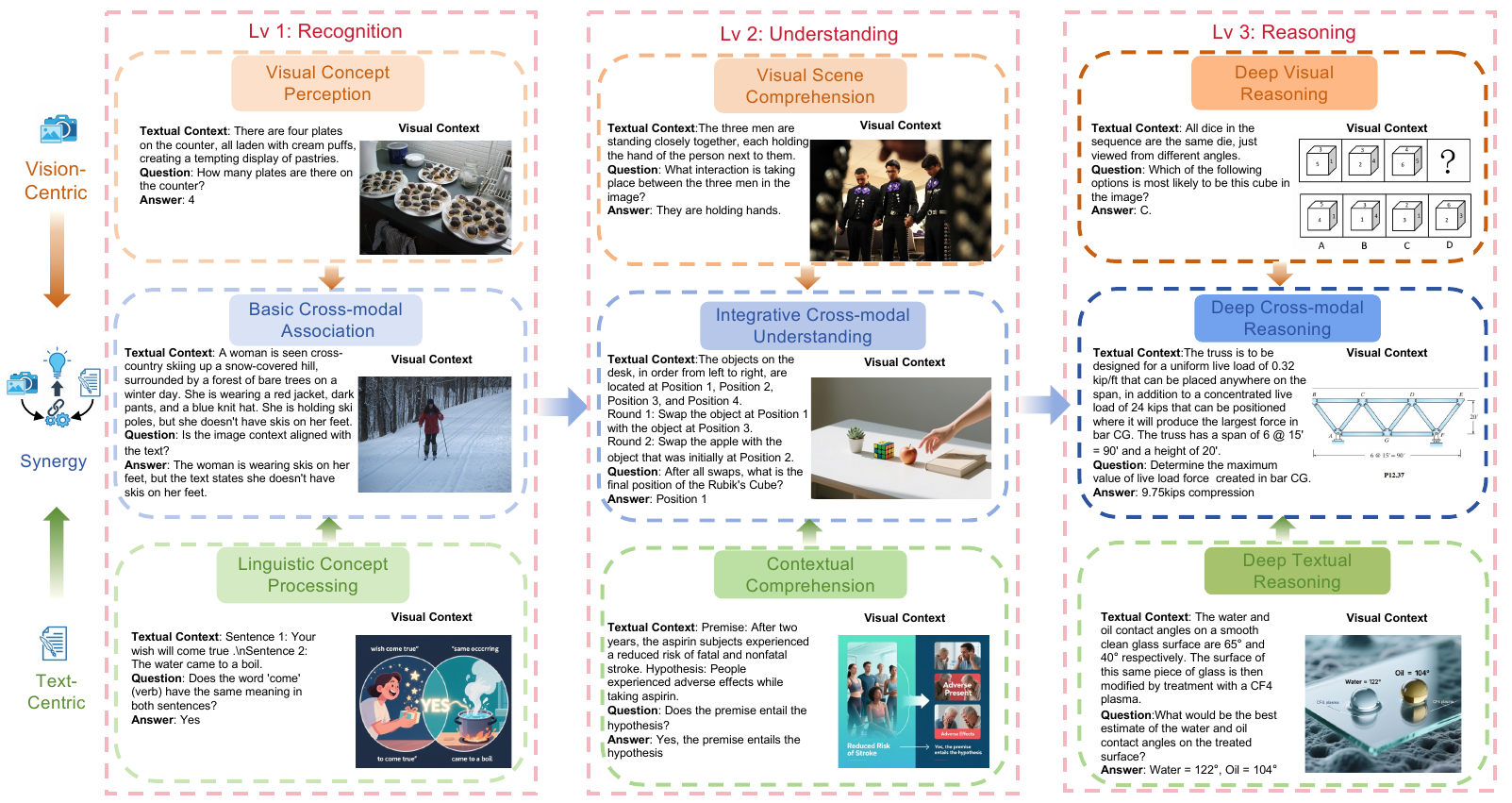}
    \vspace{-1em}
    \caption{Overview of the framework and samples. MIBench covers three interaction forms (Vision-centric, Text-centric, and Synergy) and three hierarchical levels of ability: Recognition, Understanding, and Reasoning. Each sample is composed of a textual context, a visual context, and a task (Q\&A). For the vision-centric and text-centric samples, multiple types of contexts from another modality are prepared in our evaluation, though only one is shown here for illustrative purposes.}
    \label{fig:frame}
\end{figure*}

\section{Related Works}
\label{sec:relatedworks}



\subsection{Large Multimodal Models}

LMMs are widely regarded as general-purpose problem solvers for a broad range of tasks. Initially, driven by the huge success of Large Language Models (LLMs), many approaches adopted pre-trained LLMs as a backbone, connecting them with lightweight vision components to achieve impressive scalability and performance~\cite{liu2023llava, Qwen2.5-VL,chen2024internvl}. While these models demonstrate strong results, they tend to rely heavily on the text modality. As a result, when handling vision-centric tasks, they often fail to ground in the image, making errors based on language priors instead. In contrast, native LMMs train models from scratch. This approach helps mitigate these language biases and allows the model to learn deeper, more integrated interactions between images and text~\cite{wang2024emu3, chameleonteam2024chameleon}. Meanwhile,  closed-source models remain at the forefront of general vision-language understanding~\cite{openai2023gpt4v, comanici2025gemini25pushingfrontier}. Given this diverse landscape of models, a critical question remains: How effectively do these different architectures process the fundamental interactions between modalities? Our work aims to systematically investigate this capability.


\subsection{Benchmarks for LMMs}
As a key effort to integrate vision and language, LMMs have demonstrated impressive problem-solving capabilities, even if their underlying mechanisms remain under-explored. Since questions are primarily text-based, linguistic proficiency serves as the baseline for evaluation~\cite{hendrycks2020measuring}. However, a persistent challenge is determining whether these models truly go beyond text comprehension to genuinely perceive and understand visual content. Early research like MME \cite{fu2023mme} focused on foundational perceptual abilities. Subsequent benchmarks, such as MMMU \cite{yue2024mmmu} and MMBench \cite{liu2024mmbench}, aimed to evaluate how models synthesize visual and linguistic knowledge for complex reasoning. Other studies have investigated specific issues like modality bias \cite{zheng2025mllms} and textual shortcuts \cite{zhang2024mathverse}. Despite this progress, existing evaluation frameworks struggle to capture multimodal interaction. Specifically, they fail to disentangle whether the information required to answer a question stems from the image, the text, or the synergy between them. To address this, we introduce a new evaluation paradigm that decouples visual and textual contexts, allowing us to effectively assess the multimodal interaction capabilities of LMMs.

\begin{figure*}[!t]
    \centering
    \includegraphics[width=1\textwidth]{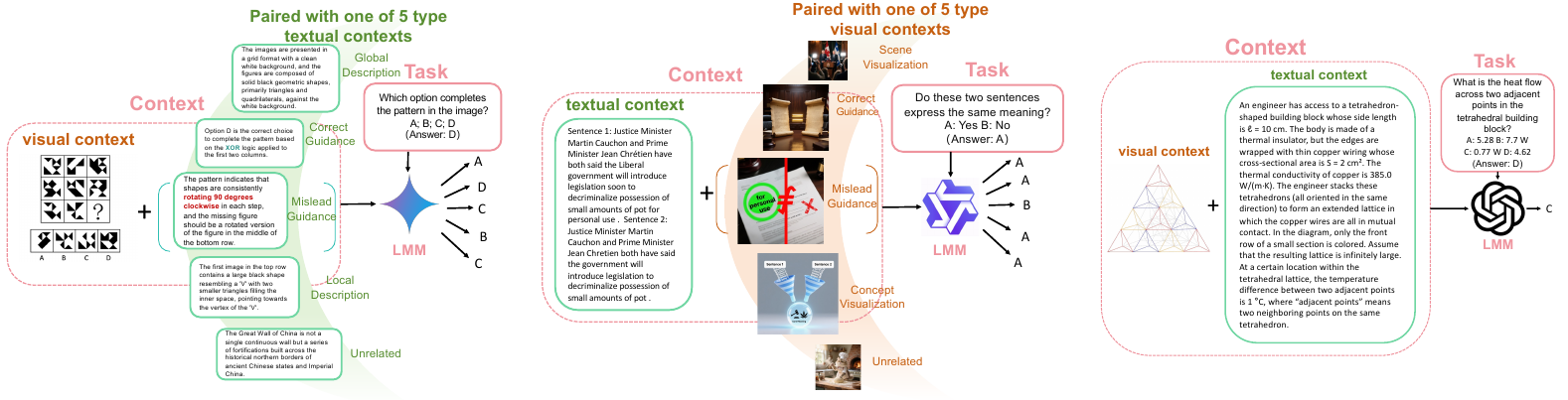}
    \vspace{-1em}
     \caption{
     Overview of MIBench sample formats.
Each instance consists of both visual and textual contexts and the corresponding task, formulated as ($con_v$, $con_t$, $task$). For vision- and text-centric tasks \textbf{(\textbf{Left} and \textbf{Middle})}, they can be resolved by leveraging cues from the centric modality. We introduce various contexts from another modality to evaluate the model's ability to selectively utilize cues from the target modality, which range from helpful contexts (e.g., correct guidance, concept visualization) to misleading guidance and unrelated contents. For the synergy part \textbf{(Right)}, the model is presented with one coupled visual-textual pair with complementary cues, necessitating effective cross-modal collaboration.
     }
     \label{fig:eva}
\end{figure*} 

\subsection{Multimodal Interaction}

Multimodal interaction describes how models collaboratively process information from different sources, such as vision and language, to solve a task~\cite{baltruvsaitis2018multimodal}. This involves two critical capabilities: the ability to source information unique to a single modality (e.g., vision or text), and the synergistic ability to integrate both modalities to derive new insights not present in either source alone. These interactions can be formally measured~\cite{liang2023quantifying, yang2025Efficient} and have been leveraged to enhance model performance~\cite{dufumier_castillo2025}.  However, while LMMs show strong performance, their complexity makes it uncertain if they use these interactions reliably~\cite{zhang2024mathverse, bai2025hallucination}. They might exploit statistical shortcuts or textual priors rather than genuinely integrating knowledge from two modalities. 

\section{The MIBench }
\label{sec:method}




In this paper, we introduce \textbf{MIBench}, a benchmark designed to evaluate LMMs on multimodal interaction. We first established a comprehensive evaluation framework from three interaction patterns as shown in Fig.~\ref{fig:frame}. 
Guided by this framework, we curate diverse tasks where each instance is formulated as a ($con_v$, $con_t$, $task$) triplet which enables assessment across different interaction mechanisms: as detailed in Fig.~\ref{fig:eva}, for vision-centric and text-centric tasks, we can vary another context to evaluate the model's proficiency in selectively extracting critical cues from the target modality; synergy tasks mandate the integration of complementary cues from both contexts to achieve cross-modal collaboration.
Consequently, MIBench stands as a comprehensive and effective benchmark for multimodal interaction evaluation, covering 32 tasks and 2,980 samples with 10,030 context pairs.

\subsection{Multimodal Interaction Framework}
\label{subsec:framework}
Fig.~\ref{fig:frame} illustrates the evaluation framework for LMMs on multimodal interaction. It forms a 3×3 grid structure based on multimodal interaction patterns and cognitive levels. In terms of interaction pattern, it consists of three parts: \textbf{Vision-centric}, \textbf{Text-centric}, and \textbf{Synergy}. 
Specifically, for vision-centric and text-centric interactions, LMMs need to selectively extract cues from the target modality; for synergistic interactions, LMMs are required to achieve synergistic collaboration between two modalities with complementary critical cues.
In terms of cognitive level, we decompose the abilities under each interaction pattern into three levels: \textbf{Recognition}, \textbf{Understanding}, and \textbf{Reasoning}. 
Concretely, this cognitive hierarchy advances from the fundamental perception of atomic concepts, through the comprehension of scenes and semantics, to the high-level reasoning of logical deduction.
Overall, this framework offer a comprehensive and fine-grained analysis of model capabilities across different multimodal interaction scenarios.


\subsection{Data Curation}
\label{subsec:datacuration}

\begin{wrapfigure}{r}{0.5\textwidth}
    \centering
    \includegraphics[width=\linewidth]{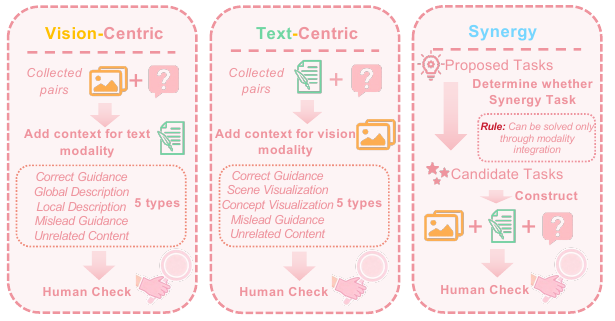}
     \caption{Overview of the sample annotation pipeline.
     }
     
    \label{fig:lalala}
\end{wrapfigure}

Guided by the framework outlined in Sec.~\ref{subsec:framework}, we constructed MIBench according to the following steps: We first identify suitable tasks based on the proposed taxonomy, and subsequently collect, re-annotate, or synthesize test samples following the annotation pipeline illustrated in Fig.~\ref{fig:lalala}. 
Finally, the instances undergo rigorous filtering and verification by human experts to ensure quality.
A more detailed description of the annotation and tasks can be found in the Appendix . 

\noindent
\textbf{Vision-centric.} We deconstruct visual-centric interaction capabilities into a three-level hierarchy: Level 1 addresses visual concept perception, involving tasks such as object recognition and counting; Level 2 pertains to visual scene comprehension, covering tasks like spatial relationships; and Level 3 focuses on deep visual reasoning, including tasks that demand the deduction of underlying rules governing visual patterns. Based on this hierarchical categorization, we first source samples from established benchmarks, including MMBench~\cite{liu2024mmbench}, MME~\cite{fu2023mme}, SeedBench~\cite{li2023seedbench2benchmarkingmultimodallarge}, and VisuLogic~\cite{xu2025visulogicbenchmarkevaluatingvisual}. Then we use Gemini 2.5 Pro~\cite{gemini2.5pro} following the pipeline illustrated in \cref{fig:lalala} to pair each visual sample with five distinct types of textual context: correct guidance, global description, local description, unrelated content, and misleading guidance. This design makes textual contexts exhibit varying degrees of relevance regarding the image and the task. Ultimately, the visual component comprises 913 base samples, with a total of 4,565 visual-textual context pairs.

\noindent
\textbf{Text-centric.} 
We stratify text-centric interaction capabilities into three progressive stages: Level 1 targets fundamental linguistic skills ranging from grammatical correctness to semantic discrimination; Level 2 pertains to broader text comprehension, exemplified by tasks like sentiment analysis; and Level 3 focuses on advanced scientific reasoning, necessitating domain-specific deduction in fields such as biology and chemistry. Guided by this stratification, we select samples from GLUE~\cite{wang2019gluemultitaskbenchmarkanalysis}, SuperGLUE~\cite{sarlin2020supergluelearningfeaturematching}, MMLU-Pro~\cite{wang2024mmluprorobustchallengingmultitask}, and GPQA~\cite{rein2023gpqagraduatelevelgoogleproofqa}. Subsequently, we employ Qwen-Image~\cite{wu2025qwenimagetechnicalreport} to associate each textual sample with five semantically distinct images: correct guidance, global description, local description, unrelated content, and misleading guidance. This design simulates diverse relationships between the visual information and the targeted textual context. Finally, the textual component comprises 1,093 base samples, yielding a total of 5,465 context pairs.


\noindent
\textbf{Synergy.}
The cross-modal synergistic capability is framed as a three-tiered hierarchy: Level 1 targets foundational alignment between two modalities, Level 2 pertains to cross-modal understanding, and Level 3 focuses on advanced reasoning using complementary cues from visual and textual contexts. 
To bridge the gap regarding foundational stages (Synergy Level 1 and 2) in prior research, we construct new tasks that enforce intensive cross-modal interaction. For Level 1, we develop tasks such as attribute mapping that require models to verify the correspondence between the fine-grained visual details and textual descriptions. For Level 2, we introduce novel tasks like rule execution, compelling the model to synthesize a comprehensive understanding by integrating textual rules with the visual scene. Images for these tasks are selected from GQA~\cite{hudson2019gqa} and COCO~\cite{lin2015microsoftcococommonobjects}, or generated by Qwen-Image, and questions are annotated by Gemini 2.5 Pro.
For Level 3, we curate samples from established multimodal reasoning benchmarks, including EMMA~\cite{hao2025mllmsreasonmultimodalityemma}, MATH-Vision~\cite{wang2024measuring}, and MMMU-Pro~\cite{yue2024mmmu}, specifically targeting those with rich textual cues, and segregate the textual contexts from their original questions. 
Collectively, the synergy section contains 974 samples covering three progressively cognitive levels.

\begin{table*}[!t]
\centering
     \small
    \caption{Performance of state-of-the-art open and closed-source models on MIBench and MIBench-mini. The best score is bolded and the second-best is underlined, within the open-source and closed-source groups separately.}
    \label{table:overallacc}
    \begin{tabular}{@{}lcccccccccc@{}}
        \toprule
        & \multicolumn{4}{c}{MIBench} & \multicolumn{4}{c}{MIBench-mini} \\
        \cmidrule(lr){2-5} \cmidrule(lr){6-9}
        Model & Vision & Text & Synergy & Overall & Vision & Text & Synergy & Overall \\
        \midrule
        \multicolumn{9}{c}{\graycell{\textit{Open-source}}} \\
        \midrule
        InternVL2-8B & 53.29 & 62.53 & 45.79 & 54.23 & 49.60 & 59.00 & 45.00  & 51.20 \\
        InternVL2.5-8B & 59.61 & 65.03 & 47.43 & 57.62 & 58.80 & 63.20 & 45.00 & 55.67 \\
        InternVL3-8B & 61.93 & 65.82 & 48.05 & 58.82 & 59.20 & 65.20 & 45.00 & 56.47 \\
        InternVL3.5-8B & 57.52 & 67.28 & 51.03 & 58.98 & 57.60 & 67.00 & 53.00 & 59.20 \\
        InternVL3.5-241B-A28B & 63.42 & \textbf{74.78} & \underline{58.01} & \underline{65.82} & 62.20 & \textbf{73.20} & \textbf{61.00} & \underline{65.47} \\
        \hdashline
        LLaVA-1.5-7B & 48.73 & 44.67 & 35.93 & 43.06 & 44.60 & 47.20 & 32.00 & 41.27 \\
        LLaVA-1.6-7B & 47.00 & 46.72 & 33.16 & 42.37 & 43.60 & 43.00 & 35.00 & 40.53 \\
        LLaVA-onevision-7B & 55.50 & 55.54 & 42.81 & 51.37 & 53.00 & 57.20 & 43.00 & 51.07 \\
        LLaVA-onevision-72B & 54.61 & 66.01 & 50.10 & 57.33 & 54.80 & 68.40 & 45.00 & 56.07 \\
        LLaVA-onevision-1.5-8B & 59.41 & 65.38 & 47.64 & 57.75 & 58.00 & 65.20 & 48.00 & 57.07 \\
        \hdashline
        Qwen2-VL-7B &60.29 & 62.76 & 44.66 & 56.09 & 59.40 & 63.80 & 43.00 & 55.40 \\
        Qwen2.5-VL-3B & 54.67 & 56.01 & 37.78 & 49.65 & 53.60 & 53.00 & 41.00 & 49.20 \\
        Qwen2.5-VL-7B & 59.61 & 58.88 & 44.97 & 54.56 & 55.80 & 59.20 & 49.00 & 54.67 \\
        Qwen2.5-VL-32B & 63.55 & 68.09 & 51.33 & 61.22 & 62.80 & 69.20 & 50.00 & 60.67 \\
        Qwen2.5-VL-72B & \underline{65.15} & 69.48 & 56.37 & 63.87 & \underline{65.40} & 67.20 & 59.00 & 63.87 \\
        Qwen3-VL-8B & 64.08 & 64.61 & 53.59 & 60.85 & 63.20 & 60.40 & 52.00 & 58.53 \\
        Qwen3-VL-235B-A22B & \textbf{68.48} & \underline{73.17} & \textbf{61.40} & \textbf{67.89} & \textbf{68.00} & \underline{72.00} & \underline{60.00} & \textbf{66.67} \\
        \hdashline
        Deepseek-VL-7B & 54.16 & 56.60 & 35.22 & 48.86 & 50.60 & 57.80 & 37.00 & 48.47 \\
        Deepseek-VL2-27B  & 61.32 & 60.49 & 40.76 & 54.30 & 58.40 & 61.20 & 39.00 & 52.87 \\
         \hdashline
        Emu3-8B  & 40.35 & 47.46 & 30.49 & 39.74 & 45.40 & 43.60 & 25.00 & 38.00 \\
        Chameleon-7B  & 29.12 & 39.58 & 37.06 & 35.55 & 24.80 & 43.00 & 34.00 & 33.93 \\
        Chameleon-30B  & 33.71 & 44.26 & 34.80 & 37.94 & 30.00 & 39.00 & 45.00 & 38.00 \\
        \midrule
        \multicolumn{9}{c}{\graycell{\textit{Closed-source}}} \\
        \midrule
        Gemini-3-pro-preview & -- & -- & -- & -- & \textbf{71.70} & \textbf{87.20} & \textbf{81.00} & \textbf{79.97} \\
        Gemini-2.5-pro & -- & -- & -- & -- & \underline{69.40} & 84.40 & \underline{76.00} & \underline{76.60} \\
        GPT-5.2 & -- & -- & -- & -- & 64.00 & 80.00 & 70.00 & 71.60 \\
        GPT-5.1 & -- & -- & -- & -- & 64.40 & 82.80 & 71.00 & 72.73 \\
        o3 & -- & -- & -- & -- & 63.60 & \underline{85.00} & 72.00 & 73.53 \\
        GPT-4o & -- & -- & -- & -- & 57.40 & 72.20 & 57.00 & 62.20 \\
        Claude-4.5-Sonnet & -- & -- & -- & -- & 59.60 & 80.80 & 72.00 & 70.80 \\
        \bottomrule
    \end{tabular}
    \vspace{-0.5em}
\end{table*}
\section{Experiments}
\label{sec:exp}
To conduct a comprehensive analysis, we assess a wide range of state-of-the-art LMMs, covering various architectures, series, versions, and parameter scales.
For open-source models, we include LLM-based architectures such as InternVL (v2.5-v3.5, 8B\&241B)~\cite{chen2024expanding,chen2024internvl, wang2025internvl3_5}, LLaVA (v1.5-OneVison1.5, 7B-72B)~\cite{liu2023improved,li2024llavaonevisioneasyvisualtask,an2025llavaonevision15fullyopenframework}, Qwen-VL (v2-v3, 3B-235B)~\cite{Qwen2VL,Qwen2.5-VL,qwen3technicalreport}, and DeepSeek-VL (v1-v2, 7B\&27B)~\cite{lu2024deepseekvl,wu2024deepseekvl2mixtureofexpertsvisionlanguagemodels} series, as well as native multimodal models like Emu3~\cite{wang2024emu3} and Chameleon (7B\&30B)~\cite{chameleonteam2024chameleon}.
Among closed-source models, we select top-performing representatives including GPT-5.1~\cite{openai-gpt5.1}, GPT-5.2~\cite{openai-gpt5.2}, GPT-4o~\cite{hurst2024gpt}, and o3~\cite{openaio3} from OpenAI, Gemini 2.5 Pro~\cite{gemini2.5pro} and Gemini 3 Pro Preview~\cite{gemini3pro} from Google, and Claude 4.5 Sonnet~\cite{anthropic_claude45_sonnet_2025} from Anthropic; and for these closed-source models, we report their performance on MIBench-mini, which is a more challenging and balanced subset carefully selected from the original benchmark, containing 100 samples for each interaction pattern.

\subsection{Evaluation Metrics}
\label{subsec:Evaluation_Metrics}

We denote the entire benchmark comprising $N$ samples as $\mathcal{D} = \{(x_i, a_i)\}_{i=1}^N$, where $x_i$ and $a_i$ represent the model input (contexts and question) of the $i$-th sample and its ground truth, respectively. The benchmark $\mathcal{D}$ is partitioned into three distinct subsets: $\mathcal{D}_{\text{vis}}$, $\mathcal{D}_{\text{txt}}$, and $\mathcal{D}_{\text{syn}}$, each targeting a specific capability: Vision-centric, Text-centric, and Synergy. We define one input unit for inference as a triple $k = (c_v, c_t, q)$, comprising the visual context, the textual context, and the question, respectively.
Accordingly, the input $x_i$ of sample in each subset can be defined as a set $\{k_{i, j}\}_{j=1}^m$, where: 
$k_{i,j}=(c_{v_i}, c_{t_{i, j}}, q_i)$ in $\mathcal{D}_{\text{vis}}$, containing a fixed visual context and a textual context varying across $m$ types; 
$k_{i,j}=(c_{v_{i,j}}, c_{t_i}, q_i)$ in $\mathcal{D}_{\text{txt}}$, containing a fixed textual context and a visual context varying across $m$ types; 
and in $\mathcal{D}_{\text{syn}}$, where both modalities provide essential information, the input is denoted simply as $x_i = \{k_i\}$.


\noindent
\textbf{Accuracy.}
We first compute a sample-level correctness score, $C(x_i)$. For a synergy sample $x_i \in D_{\text{syn}}$, which consists of only one triple, correctness is a binary value: 
$
C(x_i) = \mathbb{I}(\text{predict}(k_i) = a_i)$, where $\mathbb{I}(\cdot)$ is the indicator function. For the $i$-th sample from $D_{\text{vis}}$ or $D_{\text{txt}}$ with $m$ input variations, the score is:
\begin{equation}
C(x_i) = \frac{1}{m} \sum_{j=1}^m \mathbb{I}(\text{predict}(k_{i,j}) = a_i).
\end{equation}
The accuracy for $\mathcal{D}$ and its subsets is calculated as the average of $C(x_i)$ over all samples within each respective set.

\noindent
\textbf{Consistency.}
To specifically evaluate the model's ability to select information from centric modality, we introduce a metric called \textit{Consistency}, which is designed exclusively on $D_{\text{vis}}$ and $D_{\text{txt}}$. The responses for sample $x_i$ are considered consistent if the model gives the exact same choice for all $m$ input variations. We formalize this with a binary sample-level \textit{Consistency} score:
\begin{equation}
\label{eq:consistency_sample}
\text{Cons}(x_i) = \mathbb{I}(\left|\left\{\text{predict}(k_{i,j})\}_{j=1}^m\right| = 1\right),
\end{equation}
where $|\cdot|$ denotes the cardinality of the result set. And the overall \textit{Consistency} for the given subset ($D' \in \{D_{\text{vis}}, D_{\text{txt}}\}$) is the proportion of consistent samples, or equivalently, the average value of $\text{Cons}(x_i)$ over the entire subset.


\subsection{Limitations in Multimodal Interaction}
\label{subsec:overall}

Tab.~\ref{table:overallacc} provides an overview of current LMM performance across three interaction patterns. On the full benchmark, most models achieve accuracies between 50\% and 65\% on vision-centric tasks. Performance is generally higher on text-centric tasks, where the majority of models cluster around 65\%, with only the largest-scale models, such as InternVL3.5-241B and Qwen3-VL-235B, being capable of approaching 70\%. These results suggest that existing LMMs still struggle with insufficient targeted information extraction, particularly in the visual domain. Their performance is even lower on synergy, where most models achieve approximately 40\% accuracy. This indicates that integrating multiple information sources remains a significant challenge. \textit{Overall, most current LMMs require substantial improvement to handle comprehensive multimodal interactions.}

\begin{wrapfigure}{r}{0.5\textwidth}
    \centering
    \includegraphics[width=\linewidth]{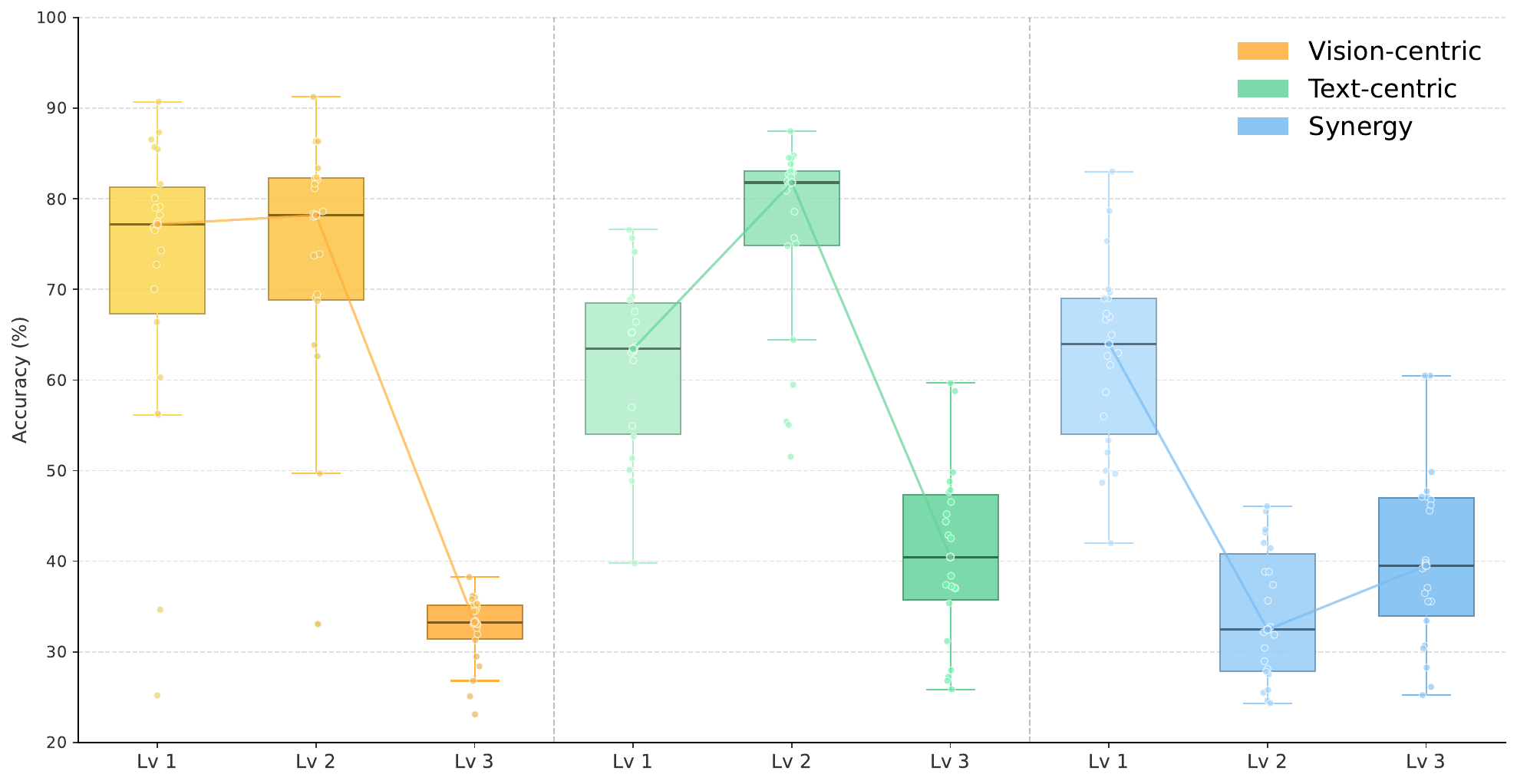}
    \caption{Performance of open-source LMMs across three progressively cognitive levels: Recognition (Level 1), Understanding (Level 2), and Reasoning (Level 3). }
    \label{fig:level}
\end{wrapfigure}
Further, Fig.~\ref{fig:level} offers a finer-grained visualization of model performance across different cognitive levels, revealing a more detailed picture of LMMs' capabilities and shortcomings. On both vision-centric and text-centric tasks, LMMs perform well on Level 1 and 2 but show a considerable decline on Level 3, which demands more complex reasoning. This suggests that while current LMMs have established basic recognition and understanding capabilities, they still lack the advanced reasoning required for sophisticated tasks. Notably, their performance on the Synergy Level 2, which stems from the understanding ability of each modality and focuses on cross-modal integration, remains poor. This indicates that while models excel at understanding information within individual modalities, they struggle to achieve deeper cross-modal synergistic emergence. To sum up, existing LMMs have demonstrated \textit{insufficient reasoning capability} and limited ability to \textit{collaboratively leverage multimodal information} for cross-modal understanding.


\begin{wrapfigure}{r}{0.5\textwidth}
    \centering
    \includegraphics[width=1\linewidth]{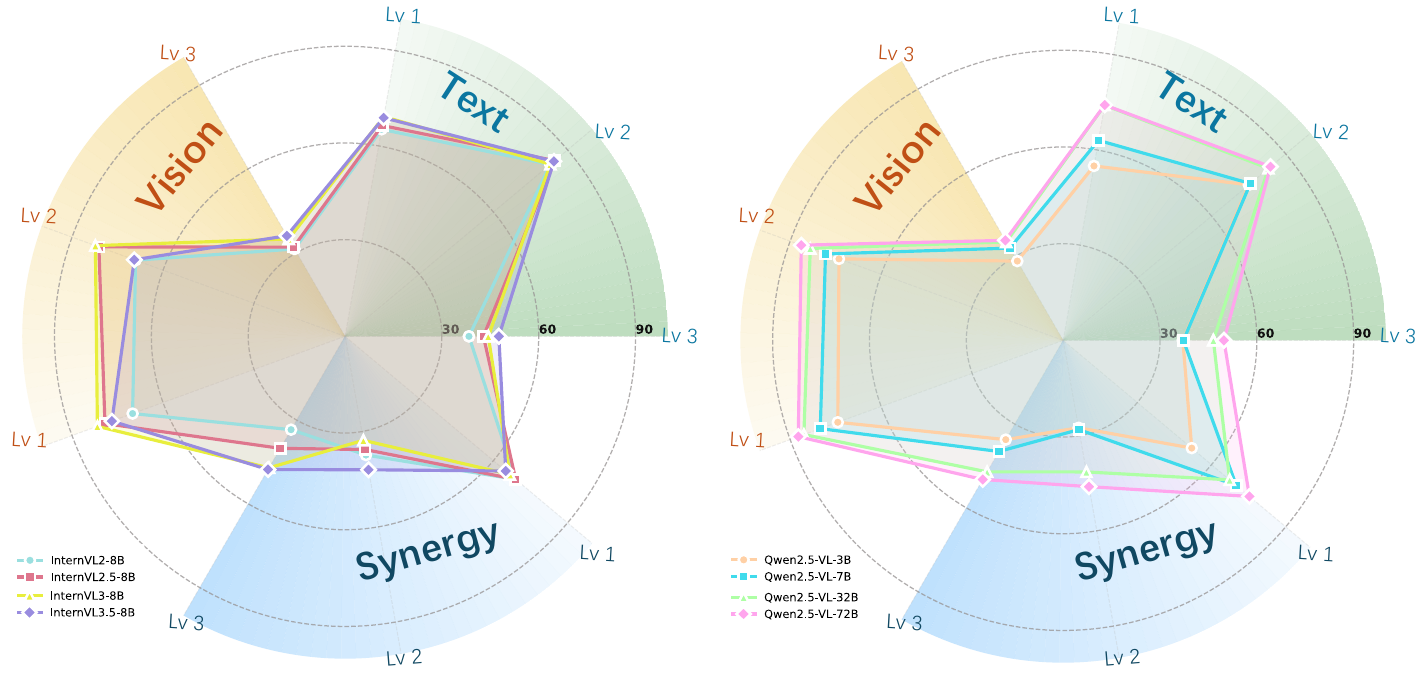}
    \caption{Impact of version iteration (\textbf{Left}) and parameter scaling (\textbf{Right}) on multimodal interaction capabilities. }
    \label{fig:compare}
\end{wrapfigure}
Under current architectures and training paradigms, both version iterations (including changes in data sources) and parameter scaling yield negligible improvements in multimodal interaction capabilities. As shown in Fig.~\ref{fig:compare} , taking the InternVL series as a case study, version updates result in minimal performance variations and, in some instances, even lead to performance regression in specific interaction phases. For example, InternVL3 and InternVL3.5 exhibit performance declines in Vision-centric Level 2 and Synergy Level 2, respectively, compared to their predecessors. This indicates that \textit{existing open-source models possess intrinsic limitations within their architectures and training paradigms}; 
Furthermore, an analysis of the QwenVL2.5 series across varying model sizes reveals the diminishing returns of scaling. While performance improves from 7B to 32B, the plateau observed between 32B and 72B indicates a saturation point, underscoring that \textit{simply expanding parameter scale is insufficient for driving substantial growth in multimodal interaction capabilities}.

Notably, most closed-source models achieve 60-75\% on MIBench-mini, surpassing the state-of-the-art open-source models by nearly 10\%. In particular, Gemini 3 Pro Preview comprehensively outperforms other proprietary models, demonstrating remarkable capabilities. 
However, this performance gap mainly stems from the text and synergy domains, where Gemini 3 Pro Preview surpasses open-source models by over 10\%. In contrast, its lead in the visual modality is relatively marginal at less than 4\%.
We hypothesize that the enhanced synergy observed in closed-source models stems primarily from their superior textual capabilities. Consequently, these results suggest that the vision-centric interaction capability of closed-source models may be the bottleneck hindering further improvements in their deeper synergy between image and texts.


\subsection{Targeted Modality Source Selection}
\label{subsec:informationselect}

\begin{wrapfigure}{r}{0.5\textwidth}
    \centering
    \includegraphics[width=1\linewidth]{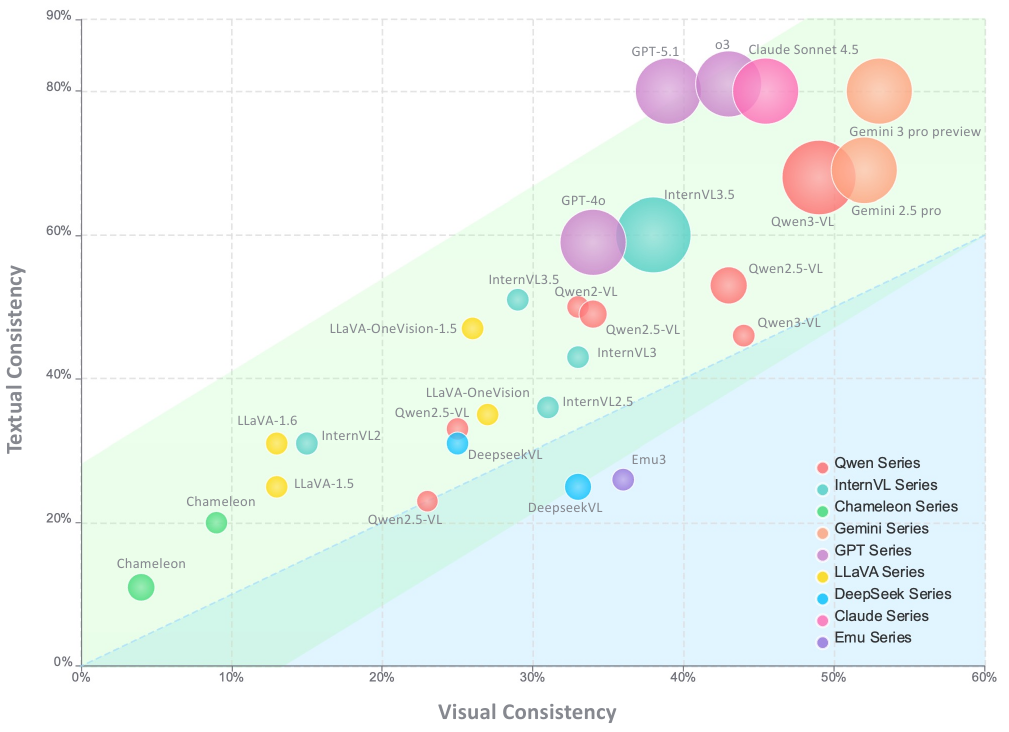}
    \caption{A comparison of LMMs' modality source selection consistency evaluated on MIBench-mini, with dot sizes scaled according to model parameter counts.}
    \label{fig:modalselection}
        \vspace{-1.5em}
\end{wrapfigure}
Based on our specialized design of multimodal contexts for vision-centric and text-centric tasks in Sec.~\ref{subsec:datacuration}, we examine whether LMMs can accurately source cues in targeted modality to handle the final task. As shown in Fig.~\ref{fig:modalselection}, the x-axis and y-axis quantify the model's \textit{Consistency} scores on vision-centric and text-centric tasks, respectively. A higher value indicates greater stability in selecting the correct information source despite variations in another modality. Ideally, LMMs should cluster in the top-right corner. However, for most open-source models, \textit{Consistency} scores across both modalities generally fall below 60\% or even 50\%, indicating a weak capability for targeted information source selection. For closed-source models, they demonstrate superior stability on sourcing textual contexts for text-centric tasks, reaching approximately 80\%, whereas their stability on selecting the targeted vision cues remains around 50\%.

Furthermore, there is a quite interesting pattern: the majority of models lie significantly above the $y = x$ line. 
This reveals a distinct textual bias: LMMs tend to prioritize text source, even in vision-centric tasks where the ground truth hinges exclusively on visual cues.
Although updates to model series and increases in parameter scales facilitate improvements in \textit{Consistency} scores of two modalities, they do not eliminate this sensitivity to variations in the textual context.
Collectively, these findings suggest that current LMMs exhibit a \textit{relatively poor ability to select the targeted modality source}, with \textit{visual information selection being particularly susceptible to interference from text source}. We attribute this bias to the fact that most existing models are built upon powerful LLM backbones, which inherently prioritize text-based representations over visual inputs.



\subsection{Synergy Deficits and Scaling Inefficiency}
\label{subsec:synergy}
As discussed in Sec.~\ref{subsec:framework}, synergy is a critical factor in assessing the capabilities of LMMs, revealing their inherent ability to achieve cross-modal collaboration. Fig.~\ref{fig:level} presents a comprehensive analysis of synergy ability across different cognition levels.
At Level 1 tasks, the median accuracy of the models exceeds 60\%, indicating that modern LMMs possess a basic ability for cross-modal alignment. This forms a foundation for more complex multimodal interactions in higher-level tasks. However, as task complexity increases to Level 2 and 3, which require reasoning capabilities in both natural and science-related contexts, model performance drops significantly. Interestingly, although Level 2 tasks primarily involve reasoning based on natural contexts, which do not need highly specialized knowledge but require frequent cross-modal interaction, the models perform even worse compared to science-related Level 3 tasks. This reveals that, \textit{while basic alignment is established, current LMMs lack the synergy ability of deep cross-modal understanding and reasoning, and may often rely on parametric knowledge to solve scientific tasks while failing at intensive cross-modal interaction in natural scenes.}

\begin{wrapfigure}{r}{0.5\textwidth}
    \centering
    \includegraphics[width=1\linewidth]{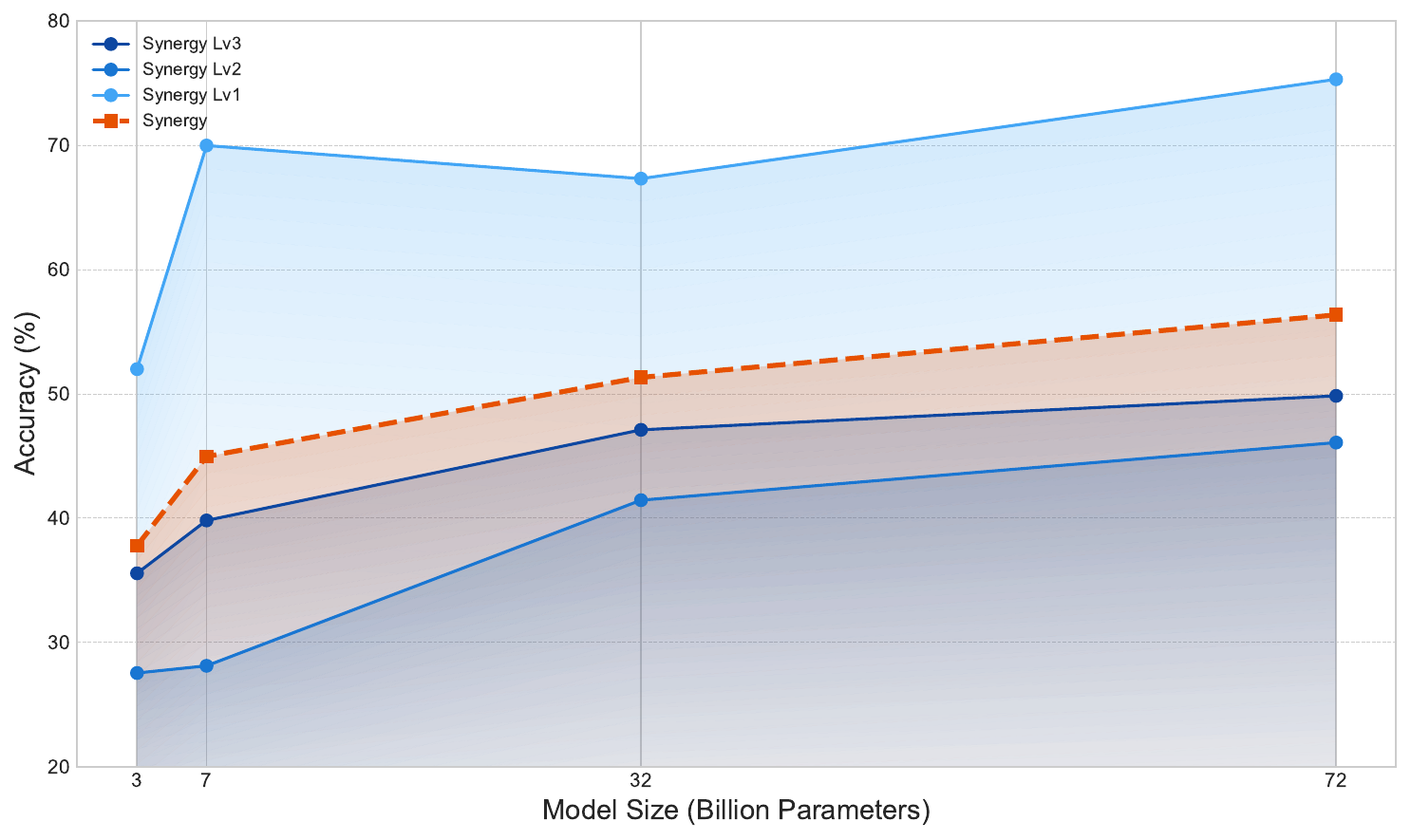}
    \caption{Impact of parameter scaling on synergy capabilities (on Qwen2.5-VL). }
    \label{fig:synergy}
    \vspace{-1.5em}
\end{wrapfigure}
Fig.~\ref{fig:synergy} illustrates the impact of parameter scaling on synergy capabilities. While a general positive correlation is observed between model size and overall accuracy, more complex interactions remain a persistent bottleneck. Despite scaling the model size to 72B, accuracies for higher-level synergy tasks (Level 2 and 3) remain consistently low (below 50\%) and exhibit a much slower growth rate compared to the massive increase in parameters. What's more, the benefits of scaling are disproportionate in the early stages. Interestingly, from 3B to 7B, the performance growth for basic tasks (Level 1) is significantly steeper than that for higher-level tasks, implying that initial scaling primarily benefits superficial interaction abilities rather than deep reasoning. Third, the gains are uneven: performance on Level 1 shows a surprising stagnation (though minor) between 7B and 32B. Overall, these results suggest that s\textit{implying scaling parameters is insufficient to overcome the intrinsic deficits in achieving deeper cross-modal collaboration}.

\subsection{Different Paradigms: Native vs. Non-native}
\label{subsec:llavaomni}
While most current LMMs are building upon a pretrained LLM backbone, like LLaVA~\cite{liu2023improved}, emerging native LMMs are trained from scratch using multimodal data, represented by Emu3~\cite{wang2024emu3} and Chameleon~\cite{chameleonteam2024chameleon}. Our experiments provide a comprehensive analysis of these two training paradigms.

As shown in Tab.~\ref{table:overallacc}, native LMMs perform poorly on both unimodal and multimodal synergy tasks, struggling even with the simplest tasks in Level 1. Under the same 7B parameter setting, Chameleon-7B shows a substantial performance gap compared with other non-native LMM counterparts, such as LLaVA-1.5-7B, Qwen2-VL-7B, and DeepSeek-VL-7B. The same phenomenon can be observed in the 8B parameter setting: Emu3-8B still falls far behind the performance of InternVL2-8B, LLaVA-OneVision-1.5-8B, and Qwen3-VL-8B. These observations suggests that, compared with non-native LMMs, these native LMMs fail to establish basic interaction capabilities, possibly because enabling cross-modal interaction at an early stage inevitably undermines unimodal performance.
Consequently, \textit{native LMMs struggle with fundamental perception, creating a bottleneck for complex interactions}, while the performance improvements of non-native LMMs \textit{can be largely attributed to advancements in the textual capabilities of the underlying pretrained LLM foundations rather than genuine multimodal synergy}.

\section{Discussion}
\label{sec:discussion}

In this work, we introduced MIBench, a benchmark that evaluates current LMMs through a structured analysis of multimodal interaction, by organizing tasks along three interaction patterns and three cognitive levels. It provides a fine-grained view of how models integrate and utilize information across modalities. Based on our evaluation, most current LMMs perform well on single-modal centric tasks or simple cross-modal tasks but degrade substantially when deeper synergistic collaboration is required. Meanwhile, LMMs built upon large language models exhibit a clear textual bias, hindering the effective utilization of visual information and synergistic information across modalities. These findings highlight the need for future LMM paradigms to achieve a better balance in modality information utilization, especially visual information. Placing greater focus on effectively leveraging each modality is essential to support diverse multimodal interactions, instead of relying predominantly on textual cues. Overall, MIBench could serve as both an evaluation tool for LMMs and a reference for developing future LMMs with more balanced modality capabilities and improved cross-modal collaboration.
\section{Acknowledgements}
This work was supported by Beijing Natural Science Foundation (4262050) and the fund for building world-class universities (disciplines) of Renmin University of China.

\clearpage

\bibliographystyle{IEEEtran}
\bibliography{paper}

@String(ICLR = {Int. Conf. Learn. Represent.})

@String(ICLR  = {ICLR})

@article{baltruvsaitis2018multimodal,
  title={Multimodal machine learning: A survey and taxonomy},
  author={Baltru{\v{s}}aitis, Tadas and Ahuja, Chaitanya and Morency, Louis-Philippe},
  journal={IEEE transactions on pattern analysis and machine intelligence},
  volume={41},
  number={2},
  pages={423--443},
  year={2018},
  publisher={IEEE}
}

@article{liang2023quantifying,
  title={Quantifying \& modeling multimodal interactions: An information decomposition framework},
  author={Liang, Paul Pu and Cheng, Yun and Fan, Xiang and Ling, Chun Kai and Nie, Suzanne and Chen, Richard and Deng, Zihao and Allen, Nicholas and Auerbach, Randy and Mahmood, Faisal and others},
  journal={Advances in Neural Information Processing Systems},
  volume={36},
  pages={27351--27393},
  year={2023}
}

@inproceedings{yang2025Efficient,
  title={Efficient Quantification of Multimodal Interaction at Sample Level},
  author={Yang, Zequn and Wang, Hongfa and Hu, Di},
  booktitle={Forty-Second International Conference on Machine Learning},
  year={2025}
}

@article{zhang2024mathverse,
  title={MathVerse: Does Your Multi-modal LLM Truly See the Diagrams in Visual Math Problems?},
  author={Zhang, Renrui and Jiang, Dongzhi and Zhang, Yichi and Lin, Haokun and Guo, Ziyu and Qiu, Pengshuo and Zhou, Aojun and Lu, Pan and Chang, Kai-Wei and Gao, Peng and others},
  journal={arXiv preprint arXiv:2403.14624},
  year={2024}
}

@inproceedings{bai2025hallucination,
    title={Hallucination at a Glance: Controlled Visual Edits and Fine-Grained Multimodal Learning},
    author={Tianyi Bai and Yuxuan Fan and Qiu Jiantao and Fupeng Sun and Jiayi Song and Junlin Han and Zichen Liu and Conghui He and Wentao Zhang and Binhang Yuan},
    booktitle={The Thirty-ninth Annual Conference on Neural Information Processing Systems},
    year={2025}
}

@article{hendrycks2020measuring,
  title={Measuring massive multitask language understanding},
  author={Hendrycks, Dan and Burns, Collin and Basart, Steven and Zou, Andy and Mazeika, Mantas and Song, Dawn and Steinhardt, Jacob},
  journal={arXiv preprint arXiv:2009.03300},
  year={2020}
}

@inproceedings{liu2024mmbench,
  title={Mmbench: Is your multi-modal model an all-around player?},
  author={Liu, Yuan and Duan, Haodong and Zhang, Yuanhan and Li, Bo and Zhang, Songyang and Zhao, Wangbo and Yuan, Yike and Wang, Jiaqi and He, Conghui and Liu, Ziwei and others},
  booktitle={European conference on computer vision},
  pages={216--233},
  year={2024},
  organization={Springer}
}

@article{fu2023mme,
  title={MME: A Comprehensive Evaluation Benchmark for Multimodal Large Language Models},
  author={Fu, Chaoyou and Chen, Peixian and Shen, Yunhang and Qin, Yulei and Zhang, Mengdan and Lin, Xu and Yang, Jinrui and Zheng, Xiawu and Li, Ke and Sun, Xing and others},
  journal={arXiv preprint arXiv:2306.13394},
  year={2023}
}

@article{zheng2025mllms,
  title={MLLMs are Deeply Affected by Modality Bias},
  author={Zheng, Xu and Liao, Chenfei and Fu, Yuqian and Lei, Kaiyu and Lyu, Yuanhuiyi and Jiang, Lutao and Ren, Bin and Chen, Jialei and Wang, Jiawen and Li, Chengxin and others},
  journal={arXiv preprint arXiv:2505.18657},
  year={2025}
}

@article{Qwen2.5-VL,
  title={Qwen2.5-VL Technical Report},
  author={Bai, Shuai and Chen, Keqin and Liu, Xuejing and Wang, Jialin and Ge, Wenbin and Song, Sibo and Dang, Kai and Wang, Peng and Wang, Shijie and Tang, Jun and Zhong, Humen and Zhu, Yuanzhi and Yang, Mingkun and Li, Zhaohai and Wan, Jianqiang and Wang, Pengfei and Ding, Wei and Fu, Zheren and Xu, Yiheng and Ye, Jiabo and Zhang, Xi and Xie, Tianbao and Cheng, Zesen and Zhang, Hang and Yang, Zhibo and Xu, Haiyang and Lin, Junyang},
  journal={arXiv preprint arXiv:2502.13923},
  year={2025}
}

@article{wu2024deepseekvl2mixtureofexpertsvisionlanguagemodels,
      title={DeepSeek-VL2: Mixture-of-Experts Vision-Language Models for Advanced Multimodal Understanding}, 
      author={Zhiyu Wu and Xiaokang Chen and Zizheng Pan and Xingchao Liu and Wen Liu and Damai Dai and Huazuo Gao and Yiyang Ma and Chengyue Wu and Bingxuan Wang and Zhenda Xie and Yu Wu and Kai Hu and Jiawei Wang and Yaofeng Sun and Yukun Li and Yishi Piao and Kang Guan and Aixin Liu and Xin Xie and Yuxiang You and Kai Dong and Xingkai Yu and Haowei Zhang and Liang Zhao and Yisong Wang and Chong Ruan},
      year={2024},
      journal={arXiv preprint arXiv:2412.10302},
}

@misc{comanici2025gemini25pushingfrontier,
      title={Gemini 2.5: Pushing the Frontier with Advanced Reasoning, Multimodality, Long Context, and Next Generation Agentic Capabilities}, 
      author={Gheorghe Comanici and Eric Bieber and Mike Schaekermann and Ice Pasupat and Noveen Sachdeva and Inderjit Dhillon and Marcel Blistein and Ori Ram and Dan Zhang and Evan Rosen and Luke Marris and Sam Petulla and Colin Gaffney and Asaf Aharoni and Nathan Lintz and Tiago Cardal Pais and Henrik Jacobsson and others},
      year={2025},
      eprint={2507.06261},
      archivePrefix={arXiv},
      journal={arXiv preprint arXiv:2507.06261},
      primaryClass={cs.CL},
      url={https://arxiv.org/abs/2507.06261}, 
}

@misc{openai-gpt5.2,
  author = {OpenAI},
  title = {{GPT-5.2}},
  howpublished = {\url{https://platform.openai.com/docs/models/gpt-5.2}},
  year = {2025},
}

@misc{openai-gpt5.1,
  author = {OpenAI},
  title = {{GPT-5.1}},
  howpublished = {\url{https://platform.openai.com/docs/models/gpt-5.1}},
  year = {2025},
}

@inproceedings{lu2024mathvista,
  author = {Lu, Pan and Bansal, Hritik and Xia, Tony and Liu, Jiacheng and Li, Chunyuan and Hajishirzi, Hannaneh and Cheng, Hao and Chang, Kai-Wei and Galley, Michel and Gao, Jianfeng},
  title = {MathVista: Evaluating Mathematical Reasoning of Foundation Models in Visual Contexts},
  booktitle = {International Conference on Learning Representations (ICLR)},
  year = {2024}
}

@article{yue2024mmmu,
  title={MMMU-Pro: A More Robust Multi-discipline Multimodal Understanding Benchmark},
  author={Xiang Yue and Tianyu Zheng and Yuansheng Ni and Yubo Wang and Kai Zhang and Shengbang Tong and Yuxuan Sun and Botao Yu and Ge Zhang and Huan Sun and Yu Su and Wenhu Chen and Graham Neubig},
  journal={arXiv preprint arXiv:2409.02813},
  year={2024}
}

@article{hao2025mllmsreasonmultimodalityemma,
      title={Can MLLMs Reason in Multimodality? EMMA: An Enhanced MultiModal ReAsoning Benchmark}, 
      author={Yunzhuo Hao and Jiawei Gu and Huichen Will Wang and Linjie Li and Zhengyuan Yang and Lijuan Wang and Yu Cheng},
      year={2025},
      journal={arXiv preprint arXiv:2501.05444},
}

@article{chen2024we,
  title={Are We on the Right Way for Evaluating Large Vision-Language Models?},
  author={Chen, Lin and Li, Jinsong and Dong, Xiaoyi and Zhang, Pan and Zang, Yuhang and Chen, Zehui and Duan, Haodong and Wang, Jiaqi and Qiao, Yu and Lin, Dahua and others},
  journal={arXiv preprint arXiv:2403.20330},
  year={2024}
}

@article{qiao2024we,
  title={We-Math: Does Your Large Multimodal Model Achieve Human-like Mathematical Reasoning?},
  author={Qiao, Runqi and Tan, Qiuna and Dong, Guanting and Wu, Minhui and Sun, Chong and Song, Xiaoshuai and GongQue, Zhuoma and Lei, Shanglin and Wei, Zhe and Zhang, Miaoxuan and others},
  journal={arXiv preprint arXiv:2407.01284},
  year={2024}
}

@inproceedings{lu2022learn,
    title={Learn to Explain: Multimodal Reasoning via Thought Chains for Science Question Answering},
    author={Lu, Pan and Mishra, Swaroop and Xia, Tony and Qiu, Liang and Chang, Kai-Wei and Zhu, Song-Chun and Tafjord, Oyvind and Clark, Peter and Ashwin Kalyan},
    booktitle={The 36th Conference on Neural Information Processing Systems (NeurIPS)},
    year={2022}
}

@article{liu2023llava,
      title={Visual Instruction Tuning}, 
      author={Haotian Liu and Chunyuan Li and Qingyang Wu and Yong Jae Lee},
      year={2023},
      eprint={2304.08485},
      archivePrefix={arXiv},
      journal={arXiv preprint arXiv:2304.08485},
      primaryClass={cs.CV}
}

@misc{liu2023improved,
      title={Improved Baselines with Visual Instruction Tuning}, 
      author={Haotian Liu and Chunyuan Li and Yuheng Li and Yong Jae Lee},
      year={2023},
      eprint={2310.03744},
      journal={arXiv preprint arXiv:2310.03744},
      archivePrefix={arXiv},
      primaryClass={cs.CV}
}

@article{li2024llavaonevisioneasyvisualtask,
      title={LLaVA-OneVision: Easy Visual Task Transfer}, 
      author={Bo Li and Yuanhan Zhang and Dong Guo and Renrui Zhang and Feng Li and Hao Zhang and Kaichen Zhang and Yanwei Li and Ziwei Liu and Chunyuan Li},
      year={2024},
      journal={arXiv preprint arXiv:2408.03326},
}

@article{an2025llavaonevision15fullyopenframework,
      title={LLaVA-OneVision-1.5: Fully Open Framework for Democratized Multimodal Training}, 
      author={Xiang An and Yin Xie and Kaicheng Yang and Wenkang Zhang and Xiuwei Zhao and Zheng Cheng and Yirui Wang and Songcen Xu and Changrui Chen and Chunsheng Wu and Huajie Tan and Chunyuan Li and Jing Yang and Jie Yu and Xiyao Wang and Bin Qin and Yumeng Wang and Zizhen Yan and Ziyong Feng and Ziwei Liu and Bo Li and Jiankang Deng},
      year={2025},
      journal={arXiv preprint arXiv:2509.23661},
}

@article{chameleonteam2024chameleon,
      title={Chameleon: Mixed-Modal Early-Fusion Foundation Models}, 
      author={Chameleon Team},
      year={2024},
      eprint={2405.09818},
      journal={arXiv preprint arXiv:2405.09818},
      archivePrefix={arXiv},
      primaryClass={cs.CL}
}

@misc{openai2023gpt4v,
      title={GPT-4V(ision) System Card},
      author={OpenAI},
      year={2023},
      howpublished={\url{https://openai.com/research/gpt-4v-system-card}},
}

@misc{gemini2.5pro,
  author = {Google Deepmind},
  title = {Gemini 2.5 Pro},
  howpublished = {\url{https://blog.google/products-and-platforms/products/gemini/gemini-2-5-pro-latest-preview/}},
  year = {2025}
}

@misc{gemini3pro,
  author = {Google Deepmind},
  title = {Gemini 3 Pro Preview},
  howpublished = {\url{https://deepmind.google/models/gemini/pro/}},
  year = {2025}
}

@article{wu2025qwenimagetechnicalreport,
      title={Qwen-Image Technical Report}, 
      author={Chenfei Wu and Jiahao Li and Jingren Zhou and Junyang Lin and Kaiyuan Gao and Kun Yan and Sheng-ming Yin and Shuai Bai and Xiao Xu and Yilei Chen and Yuxiang Chen and Zecheng Tang and Zekai Zhang and Zhengyi Wang and An Yang and Bowen Yu and Chen Cheng and Dayiheng Liu and Deqing Li and Hang Zhang and Hao Meng and Hu Wei and Jingyuan Ni and Kai Chen and Kuan Cao and Liang Peng and Lin Qu and Minggang Wu and Peng Wang and Shuting Yu and Tingkun Wen and Wensen Feng and Xiaoxiao Xu and Yi Wang and Yichang Zhang and Yongqiang Zhu and Yujia Wu and Yuxuan Cai and Zenan Liu},
      year={2025},
      journal={arXiv preprint arXiv:2508.02324},
}

@article{li2023seedbench2benchmarkingmultimodallarge,
      title={SEED-Bench-2: Benchmarking Multimodal Large Language Models}, 
      author={Bohao Li and Yuying Ge and Yixiao Ge and Guangzhi Wang and Rui Wang and Ruimao Zhang and Ying Shan},
      year={2023},
      journal={arXiv preprint arXiv:2311.17092},
}

@article{wang2019gluemultitaskbenchmarkanalysis,
      title={GLUE: A Multi-Task Benchmark and Analysis Platform for Natural Language Understanding}, 
      author={Alex Wang and Amanpreet Singh and Julian Michael and Felix Hill and Omer Levy and Samuel R. Bowman},
      year={2019},
      journal={arXiv preprint arXiv:1804.07461},
}

@article{sarlin2020supergluelearningfeaturematching,
      title={SuperGlue: Learning Feature Matching with Graph Neural Networks}, 
      author={Paul-Edouard Sarlin and Daniel DeTone and Tomasz Malisiewicz and Andrew Rabinovich},
      year={2020},
      journal={arXiv preprint arXiv:1911.11763},
}

@article{rein2023gpqagraduatelevelgoogleproofqa,
      title={GPQA: A Graduate-Level Google-Proof Q\&A Benchmark}, 
      author={David Rein and Betty Li Hou and Asa Cooper Stickland and Jackson Petty and Richard Yuanzhe Pang and Julien Dirani and Julian Michael and Samuel R. Bowman},
      year={2023},
      journal={arXiv preprint arXiv:2311.12022},
}

@article{wang2024mmluprorobustchallengingmultitask,
      title={MMLU-Pro: A More Robust and Challenging Multi-Task Language Understanding Benchmark}, 
      author={Yubo Wang and Xueguang Ma and Ge Zhang and Yuansheng Ni and Abhranil Chandra and Shiguang Guo and Weiming Ren and Aaran Arulraj and Xuan He and Ziyan Jiang and Tianle Li and Max Ku and Kai Wang and Alex Zhuang and Rongqi Fan and Xiang Yue and Wenhu Chen},
      year={2024},
      journal={arXiv preprint arXiv:2406.01574},
}

@inproceedings{
wang2024measuring,
title={Measuring Multimodal Mathematical Reasoning with MATH-Vision Dataset},
author={Ke Wang and Junting Pan and Weikang Shi and Zimu Lu and Houxing Ren and Aojun Zhou and Mingjie Zhan and Hongsheng Li},
booktitle={The Thirty-eight Conference on Neural Information Processing Systems Datasets and Benchmarks Track},
year={2024},
}

@inproceedings{hudson2019gqa,
  title={Gqa: A new dataset for real-world visual reasoning and compositional question answering},
  author={Hudson, Drew A and Manning, Christopher D},
  booktitle={Proceedings of the IEEE/CVF conference on computer vision and pattern recognition},
  pages={6700--6709},
  year={2019}
}

@article{lin2015microsoftcococommonobjects,
      title={Microsoft COCO: Common Objects in Context}, 
      author={Tsung-Yi Lin and Michael Maire and Serge Belongie and Lubomir Bourdev and Ross Girshick and James Hays and Pietro Perona and Deva Ramanan and C. Lawrence Zitnick and Piotr Dollár},
      year={2015},
      journal={arXiv preprint arXiv:1405.0312},
}

@inproceedings{chen2024internvl,
  title={Internvl: Scaling up vision foundation models and aligning for generic visual-linguistic tasks},
  author={Chen, Zhe and Wu, Jiannan and Wang, Wenhai and Su, Weijie and Chen, Guo and Xing, Sen and Zhong, Muyan and Zhang, Qinglong and Zhu, Xizhou and Lu, Lewei and others},
  booktitle={Proceedings of the IEEE/CVF Conference on Computer Vision and Pattern Recognition},
  pages={24185--24198},
  year={2024}
}

@article{chen2024expanding,
  title={Expanding Performance Boundaries of Open-Source Multimodal Models with Model, Data, and Test-Time Scaling},
  author={Chen, Zhe and Wang, Weiyun and Cao, Yue and Liu, Yangzhou and Gao, Zhangwei and Cui, Erfei and Zhu, Jinguo and Ye, Shenglong and Tian, Hao and Liu, Zhaoyang and others},
  journal={arXiv preprint arXiv:2412.05271},
  year={2024}
}

@article{wang2025internvl3_5,
  title={InternVL3.5: Advancing Open-Source Multimodal Models in Versatility, Reasoning, and Efficiency},
  author={Wang, Weiyun and Gao, Zhangwei and Gu, Lixin and Pu, Hengjun and Cui, Long and Wei, Xingguang and Liu, Zhaoyang and Jing, Linglin and Ye, Shenglong and Shao, Jie and others},
  journal={arXiv preprint arXiv:2508.18265},
  year={2025}
}

@article{Qwen2VL,
  title={Qwen2-VL: Enhancing Vision-Language Model's Perception of the World at Any Resolution},
  author={Wang, Peng and Bai, Shuai and Tan, Sinan and Wang, Shijie and Fan, Zhihao and Bai, Jinze and Chen, Keqin and Liu, Xuejing and Wang, Jialin and Ge, Wenbin and Fan, Yang and Dang, Kai and Du, Mengfei and Ren, Xuancheng and Men, Rui and Liu, Dayiheng and Zhou, Chang and Zhou, Jingren and Lin, Junyang},
  journal={arXiv preprint arXiv:2409.12191},
  year={2024}
}

@article{lu2024deepseekvl,
      title={DeepSeek-VL: Towards Real-World Vision-Language Understanding}, 
      author={Haoyu Lu and Wen Liu and Bo Zhang and Bingxuan Wang and Kai Dong and Bo Liu and Jingxiang Sun and Tongzheng Ren and Zhuoshu Li and Yaofeng Sun and Chengqi Deng and Hanwei Xu and Zhenda Xie and Chong Ruan},
      year={2024},
      journal={arXiv preprint arXiv:2403.05525},
}

@article{wang2024emu3,
  title={Emu3: Next-Token Prediction is All You Need},
  author={Wang, Xinlong and Zhang, Xiaosong and Luo, Zhengxiong and Sun, Quan and Cui, Yufeng and Wang, Jinsheng and Zhang, Fan and Wang, Yueze and Li, Zhen and Yu, Qiying and others},
  journal={arXiv preprint arXiv:2409.18869},
  year={2024}
}

@article{qwen3technicalreport,
      title={Qwen3 Technical Report}, 
      author={Qwen Team},
      year={2025},
      journal={arXiv preprint arXiv:2505.09388},
}

@article{hurst2024gpt,
  title={Gpt-4o system card},
  author={Hurst, Aaron and Lerer, Adam and Goucher, Adam P and Perelman, Adam and Ramesh, Aditya and Clark, Aidan and Ostrow, AJ and Welihinda, Akila and Hayes, Alan and Radford, Alec and others},
  journal={arXiv preprint arXiv:2410.21276},
  year={2024}
}

@misc{openaio3,
  title={OpenAI o3 and o4-mini System Card},
  author={OpenAI},
  year={2025},
  howpublished={\url{https://openai.com/index/o3-o4-mini-system-card/}},
}

@misc{anthropic_claude45_sonnet_2025,
  author       = {{Anthropic}},
  title        = {Claude 4.5 Sonnet},
  howpublished = {\url{https://www.anthropic.com/news/claude-4-5-sonnet}},
  year         = {2025}
}

@inproceedings{dufumier_castillo2025, 
    title={What to align in multimodal contrastive learning?},
    author={Dufumier, Benoit and Castillo-Navarro, Javiera and Tuia, Devis and Thiran, Jean-Philippe},
    booktitle={International Conference on Learning Representations},
    year={2025}
}

@article{xu2025visulogicbenchmarkevaluatingvisual,
      title={VisuLogic: A Benchmark for Evaluating Visual Reasoning in Multi-modal Large Language Models}, 
      author={Weiye Xu and Jiahao Wang and Weiyun Wang and Zhe Chen and Wengang Zhou and Aijun Yang and Lewei Lu and Houqiang Li and Xiaohua Wang and Xizhou Zhu and Wenhai Wang and Jifeng Dai and Jinguo Zhu},
      year={2025},
      journal = {arXiv preprint arXiv:2504.15279},
}

\clearpage

\beginappendix









\section{Additional Results}

\subsection{Detailed Interaction Analysis}


Fig.~\ref{fig:level} in the main paper provides a holistic view of how current LMMs process multimodal interactions under different levels. To explore this further, we conducted a more fine-grained comparison of LMMs' performance based on the following aspects:


\begin{figure*}[h]
    \centering
    \includegraphics[width=0.7\textwidth]{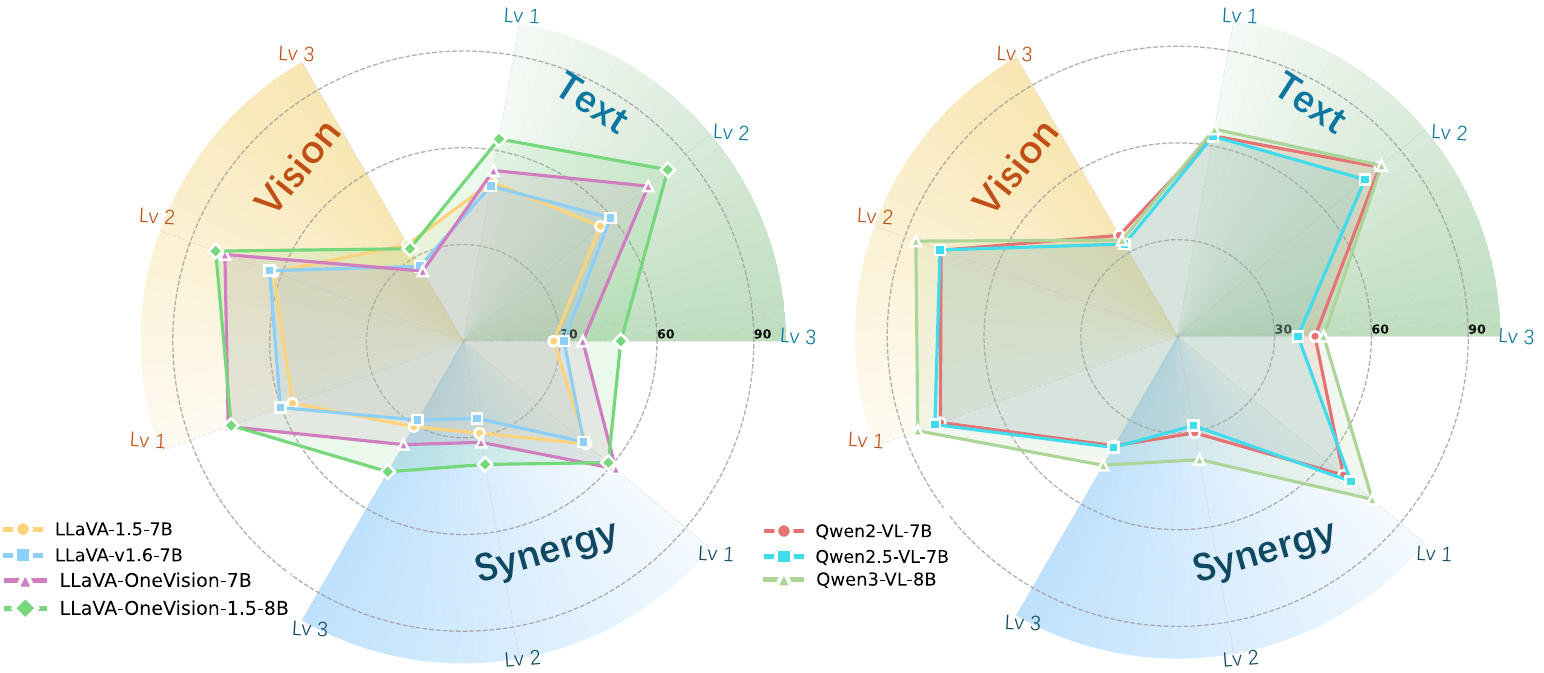}
     \caption{Comparison of LMMs' performance within different series. (\textbf{Left:} LLaVA; \textbf{Right:} Qwen-VL) 
     }
     \label{fig:1}
\end{figure*}

Firstly, the iterative evolution of model series does not necessarily yield consistent improvements in interaction capabilities across all dimensions, as illustrated in Fig.~\ref{fig:1}. For instance, within the LLaVA series, while LLaVA-OneVision-7B demonstrates significant enhancements in Vision-centric Level 1 and 2, it exhibits a regression at Vision-centric Level 3. Similarly, the Qwen2.5 series shows a degradation in Synergy Level 2 and Text Level 2. 

\begin{figure*}[h]
    \centering
    \includegraphics[width=0.65\textwidth]{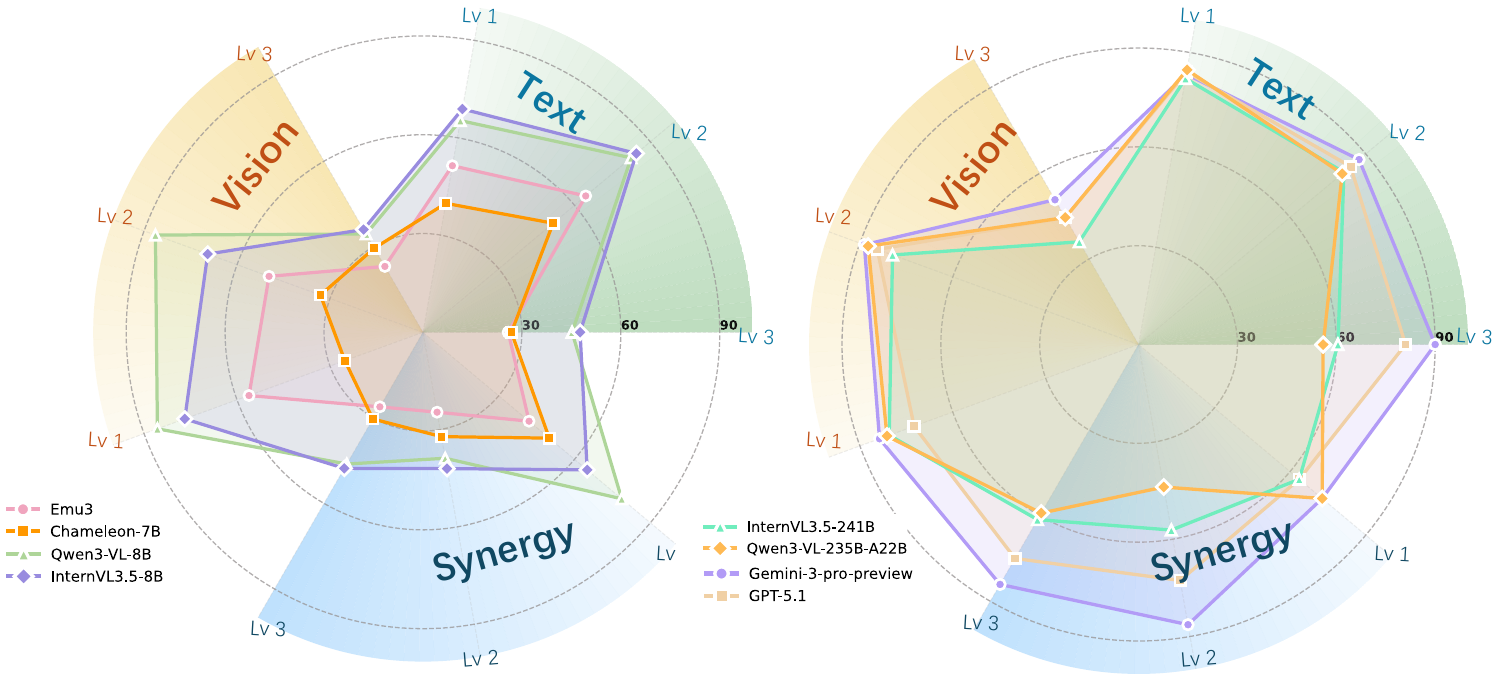}
     \caption{\textbf{Left:} non-native vs. native LMMs. \textbf{Right:} Performance of SOTA open-source and closed-source LMMs.  
     }
     \label{fig:two}
\end{figure*}

Secondly, current Native LMMs show distinct limitations. As shown in Fig.~\ref{fig:two}, they score significantly lower on Vision-centric Level 1 and Text-centric Level 1 compared to non-native models. This indicates that native multimodal models have not yet fully learned to distinguish and utilize information from specific modalities effectively. Furthermore, they do not demonstrate stable improvements in synergy capabilities, highlighting a key direction for future research.

Finally, regarding best-performing LMMs, Gemini 3 Pro Preview demonstrates remarkable SOTA performance across diverse interaction abilities as depicted in Fig.~\ref{fig:two}. Most notably, it demonstrates a substantial improvement at Synergy Level 2 compared to other models.
Among open-source models, InternVL3.5-241B and Qwen3-VL-235B demonstrate comparable performance in vision and text modalities; however, InternVL3.5-241B exhibits notably superior capabilities at Synergy Level 2.

\subsection{Task-specific interaction analysis}
The MIBench hierarchy is structured based on the specific capabilities required by each task. It categorizes these requirements into three modalities (Vision-centric, Text-centric, and Synergy) and three cognitive levels (Recognition, Understanding, and Reasoning). By analyzing model performance across 32 distinct tasks, we provide a deeper insight into LMMs' capabilities.
\begin{figure*}[h]
    \centering
    \includegraphics[width=0.9\textwidth]{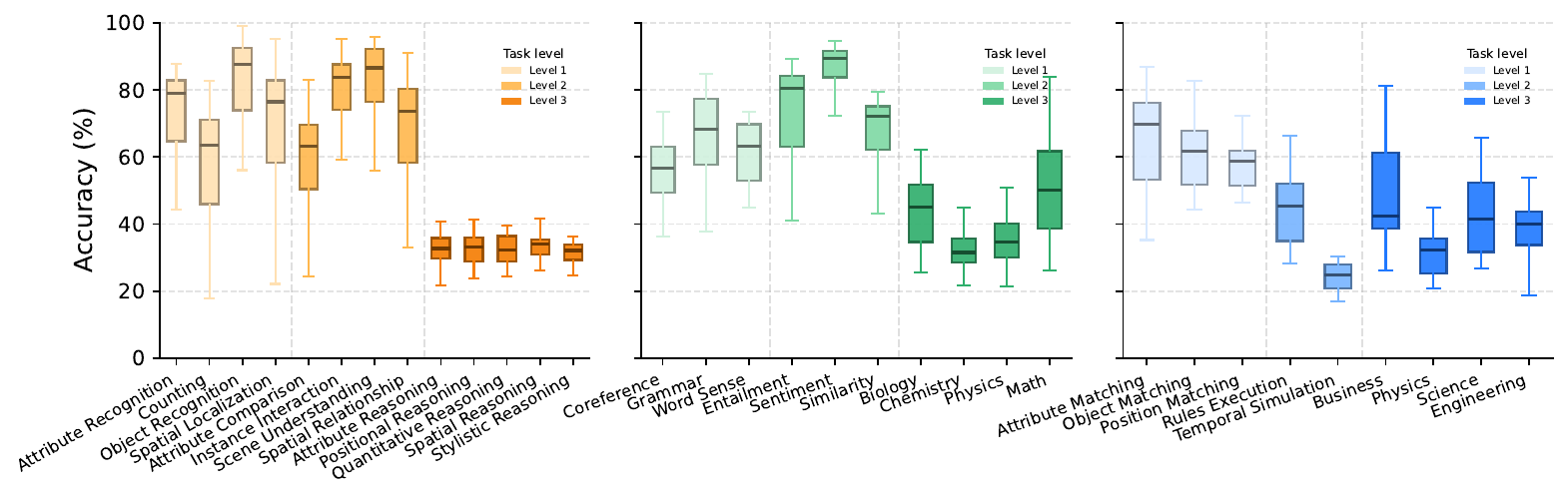}
    \vspace{-1.5em}
     \caption{LMM's performance on different task types.
     }
         \vspace{-1.5em}
     \label{fig:task}
\end{figure*}

As shown in Fig.~\ref{fig:task}, there is a clear dependency between task complexity and cognitive level. Mastery of low-level concepts, such as Attribute Recognition and Word Sense, serves as the necessary foundation for advanced understanding tasks like Spatial Relationship detection and Sentiment analysis.

In the evaluation for Vision-centric and Text-centric ability, models demonstrate relatively strong performance in Recognition and Understanding. However, a significant gap exists in Reasoning. LMMs struggle to reason based on pure visual inputs and show similar weaknesses in linguistic reasoning. The notable exceptions are Mathematics and Biology. This divergence likely occurs because mathematics is a foundational skill heavily emphasized during training, while biology is a knowledge-intensive subject where models can often rely on memorization rather than logical inference to answer questions.

Finally, regarding Synergy, performance drops specifically at Level 2, which demands complex image-text reasoning. The Temporal Simulation task yields the lowest scores, indicating that LMMs still fail to effectively synthesize information when images and text present heterogeneous data.

\section{Additional Details of MIBench}
\label{benchmarkdetails}

\subsection{Detailed Statistics}
\begin{figure*}[!h]
    \centering
    \includegraphics[width=0.8\textwidth]{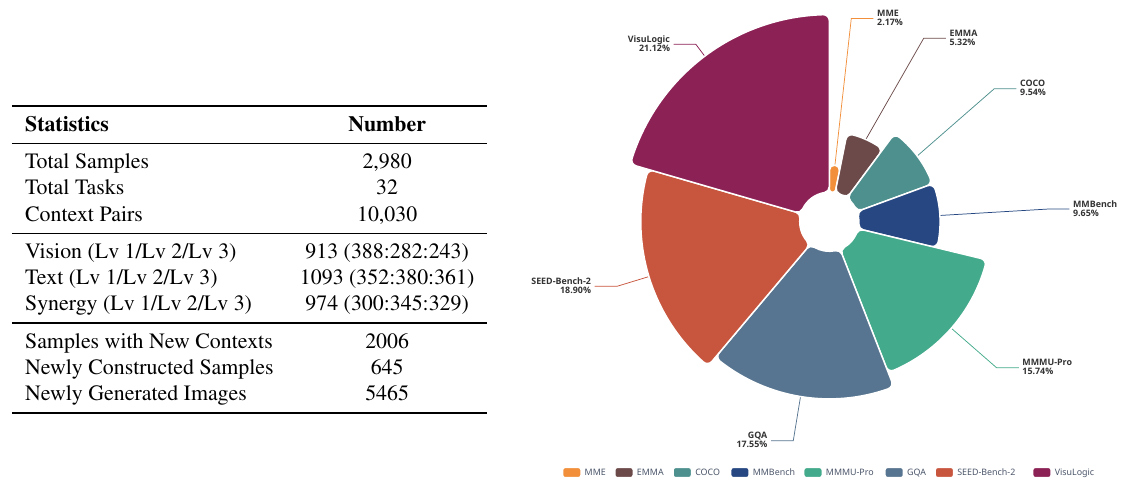}
     \caption{The left table shows more statistics of MIBench. The right graph shows proportion of sources for collected images.
     }
     \label{fig:rose}
\end{figure*}

We provide Fig.~\ref{fig:rose} with more detailed statistics for supplement.

\vspace{-1em}

\begin{table*}[htbp]
\centering
\caption{Detailed source information of the LMMs evaluated in this study.}
\label{tab:model-sources}
\small
\begin{tabularx}{\linewidth}{@{} l l >{\raggedright\arraybackslash\hsize=0.4\hsize}X >{\raggedright\arraybackslash\hsize=1.6\hsize}X @{}}
\toprule
\textbf{Model} & \textbf{Source} & \textbf{Version} & \textbf{URL} \\ 
\midrule
\multicolumn{4}{c}{\graycell{\textit{Closed-source Models}}} \\
\midrule
Gemini-3-pro-preview & Google & stable & \url{https://ai.google.dev/gemini-api/docs/models} \\
Gemini-2.5-pro & Google & stable & \url{https://ai.google.dev/gemini-api/docs/models} \\
Gemini-2.5-flash & Google & stable & \url{https://ai.google.dev/gemini-api/docs/models} \\
GPT-5.1 & OpenAI & 2025-11-13 & \url{https://platform.openai.com/docs/models/gpt-5.1} \\
GPT-5.2 & OpenAI & 2025-12-11 & \url{https://platform.openai.com/docs/models/gpt-5.2} \\
o3 & OpenAI & 2025-04-16 & \url{https://platform.openai.com/docs/models/o3} \\
GPT-4o & OpenAI & 2024-11-20 & \url{https://platform.openai.com/docs/models/gpt-4o} \\
Claude-4.5-Sonnet & Anthropic & 2025-09-29 & \url{https://www.anthropic.com/claude/sonnet} \\ 
\midrule
\multicolumn{4}{c}{\graycell{\textit{Open-source Models}}} \\
\midrule
InternVL2-8B & OpenGVLab & Latest Released & \url{https://huggingface.co/OpenGVLab/InternVL2-8B} \\
InternVL2.5-8B & OpenGVLab & Latest Released & \url{https://huggingface.co/OpenGVLab/InternVL2_5-8B} \\
InternVL3-8B & OpenGVLab & Latest Released & \url{https://huggingface.co/OpenGVLab/InternVL3-8B} \\
InternVL3.5-8B & OpenGVLab & Latest Released & \url{https://huggingface.co/OpenGVLab/InternVL3_5-8B} \\
InternVL3.5-241B-A28B & OpenGVLab & Latest Released & \url{https://huggingface.co/OpenGVLab/InternVL3_5-241B-A28B} \\
LLaVA-1.5-7B & Liuhaotian & Latest Released & \url{https://huggingface.co/liuhaotian/llava-v1.5-7b} \\
LLaVA-1.6-7B & Liuhaotian & Latest Released & \url{https://huggingface.co/liuhaotian/llava-v1.6-vicuna-7b} \\
LLaVA-onevision-7B & LMMS-Lab & Latest Released & \url{https://huggingface.co/lmms-lab/llava-onevision-qwen2-7b-ov} \\
LLaVA-onevision-72B & LMMS-Lab & Latest Released & \url{https://huggingface.co/lmms-lab/llava-onevision-qwen2-72b-ov} \\
LLaVA-onevision-1.5-8B & LMMS-Lab & Latest Released & \url{https://huggingface.co/lmms-lab/LLaVA-OneVision-1.5-8B-Instruct} \\
Qwen2-VL-7B & Alibaba & Latest Released & \url{https://huggingface.co/Qwen/Qwen2-VL-7B-Instruct} \\
Qwen2.5-VL-3B & Alibaba & Latest Released & \url{https://huggingface.co/Qwen/Qwen2.5-VL-3B-Instruct} \\
Qwen2.5-VL-7B & Alibaba & Latest Released & \url{https://huggingface.co/Qwen/Qwen2.5-VL-7B-Instruct} \\
Qwen2.5-VL-32B & Alibaba & Latest Released & \url{https://huggingface.co/Qwen/Qwen2.5-VL-32B-Instruct} \\
Qwen2.5-VL-72B & Alibaba & Latest Released & \url{https://huggingface.co/Qwen/Qwen2.5-VL-72B-Instruct} \\
Qwen3-VL-8B & Alibaba & Latest Released & \url{https://huggingface.co/Qwen/Qwen3-VL-8B-Instruct} \\
Qwen3-VL-235B-A22B & Alibaba & Latest Released & \url{https://huggingface.co/Qwen/Qwen3-VL-235B-A22B-Instruct} \\
Deepseek-VL-7B & DeepSeek-AI & Latest Released & \url{https://huggingface.co/deepseek-ai/deepseek-vl-7b-chat} \\
Deepseek-VL2-27B & DeepSeek-AI & Latest Released & \url{https://huggingface.co/deepseek-ai/deepseek-vl2} \\
Emu3-8B & BAAI & Latest Released & \url{https://huggingface.co/BAAI/Emu3-Chat} \\
Chameleon-7B & Meta & Latest Released & \url{https://huggingface.co/facebook/chameleon-7b} \\
Chameleon-30B & Meta & Latest Released & \url{https://huggingface.co/facebook/chameleon-30b} \\ 
\bottomrule
\end{tabularx}
\end{table*}

\section{Experiment Setting}
\label{expdetails}

\subsection{Evaluated Models and Computational Setup}
We evaluated 30 open-source and closed-source LMMs, with detailed specifications provided in Fig.~\ref{tab:model-sources}. For closed-source models, we accessed them via their official APIs. For open-source models, we performed inference using the Transformers library on NVIDIA H200 140GB GPUs. To manage computational resources efficiently, we allocated GPUs based on model size: a single GPU for models ranging from 3B to 32B, two GPUs for 72B models, and four GPUs for models exceeding 200B parameters.

\subsection{Prompts for Inference}
As described in Section 4, we provide the models with visual-textual contexts and questions to generate responses. Given that our questions are in a multiple-choice format, we instruct the models to output the final option letter in a fixed format to facilitate the extraction of their choices. 
\begin{tcolorbox}[mybox]
\{visual\_context\}\{textual\_context\}\{question\}\{options\}

Please ensure that your output only contains the final answer in one  `\verb|\boxed{}|' without any additional content.
\end{tcolorbox}

In particular, for models with an intrinsic thinking mode, we allow them to solve the problems step-by-step. 
\begin{tcolorbox}[mybox]
\{visual\_context\}\{textual\_context\}\{question\}\{options\}

Please solve the problem step by step, and put the option of the final answer in one
`\verb|\boxed{}|'.
\end{tcolorbox}

\subsection{Detailed Definition of Tasks}
To evaluate the capabilities of LMMs across three interaction modalities, MIBench encompasses 32 tasks spanning three levels: from recognition, understanding, to reasoning.

For the visual part, Level 1 tests the LMM’s ability to perceive visual concepts, including the following tasks:
\vspace{-1mm}
\begin{itemize}[nosep,leftmargin=*]
\item \textbf{Attribute Recognition.} Identifies and classifies visual attributes such as color, shape, size, or material, assessing the model’s ability to detect and describe distinguishing object characteristics.
\item \textbf{Counting.} Quantifies specific object instances, evaluating the ability to recognize and enumerate targets accurately.
\item \textbf{Object Recognition.} Identifies and categorizes objects within an image, testing the ability to correctly label and distinguish between different objects based on visual features.
\item \textbf{Spatial Localization.} Determines the precise location of objects within an image, requiring the model to identify exactly where specific elements are situated in the visual context.
\end{itemize}
\vspace{-1mm}
Visual Level 2 focuses on the LMM’s ability to understand and interpret visual scenes. In particular, it also includes a fine-grained comparison of the relationships between the objects contained in the scene. The tasks included are as follows:
\vspace{-1mm}
\begin{itemize}[nosep,leftmargin=*]
\item \textbf{Attribute Comparison.} Compares visual properties (e.g., size, color, shape) across multiple objects to assess differences or similarities between elements within the scene.
\item \textbf{Instance Interaction.} Analyzes dynamic relationships between objects, such as physical contact, causal effects, or specific interactions that occur between different instances.
\item \textbf{Scene Understanding.} Comprehends the holistic context, setting, and broader narrative of a visual scene, rather than focusing solely on individual isolated objects.
\item \textbf{Spatial Relationship.} Identifies the relative positioning (e.g., ``behind'', ``left of''), orientation, and spatial distribution of multiple objects across the visual context.
\end{itemize}
\vspace{-1mm}
Visual Level 3 focuses on the LMM’s visual reasoning ability to find patterns between visual images and make predictions. The tasks included are as follows:
\vspace{-1mm}
\begin{itemize}[nosep,leftmargin=*]
\item \textbf{Quantitative Reasoning.} Derives arithmetic relationships and makes predictions based on the quantification of graphical elements like lines, shapes, and points in the image.
\item \textbf{Spatial Reasoning.} Interprets 2D representations to understand 3D structures, involving tasks such as reconstruction, folding analysis, and understanding spatial assembly mechanisms.
\item \textbf{Positional Reasoning.} Predicts object states following geometric transformations(e.g., translation, rotation, and reflection)while maintaining the integrity of basic properties.
\item \textbf{Attribute Reasoning.} Deduces intrinsic geometric properties, such as symmetry, curvature, and topology, to predict how they affect the object's overall structure.
\item \textbf{Stylistic Reasoning.} Analyzes stylistic operations (e.g., overlaying, subtraction) to assess the similarity between images.
\end{itemize}
\vspace{-1mm}
For the textual part, Level 1 focuses on the LMM’s ability to comprehend fundamental grammar and word concepts. The tasks included are as follows:
\vspace{-1mm}
\begin{itemize}[nosep,leftmargin=*]
\item \textbf{Coreference.} Determines whether different linguistic expressions (e.g., pronouns, noun phrases) within the text refer to the same underlying entity or concept.
\item \textbf{Grammatical.} Validates sentence structure, syntax, and punctuation to determine if the text strictly adheres to standard grammatical rules and conventions.
\item \textbf{Word Sense.} Disambiguates and compares word meanings based on their shifting usage in different linguistic contexts to interpret the correct sense.
\end{itemize}
\vspace{-1mm}
Text Level 2 focuses on the LMM’s ability to understand and interpret textual meaning. The tasks included are as follows:
\vspace{-1mm}
\begin{itemize}[nosep,leftmargin=*]
\item \textbf{Entailment.} Determines if a premise logically guarantees the truth of a hypothesis, assessing the validity of logical implications between statements.
\item \textbf{Sentiment.} Identifies the specific emotional tone (positive or negative) expressed in a text, such as a review, opinion, or statement.
\item \textbf{Similarity.} Assesses whether two sentences share the same semantic meaning or act as paraphrases based on their content and structure.
\end{itemize}
\vspace{-1mm}
Text Level 3 focuses on the LMM’s ability to reason with text in specific scientific and mathematical domains. The tasks included are as follows:
\vspace{-1mm}
\begin{itemize}[nosep,leftmargin=*]
\item \textbf{Biology.} Reasons about biological concepts, processes, and relationships(e.g., cellular functions, genetics, and anatomy)are derived directly from textual descriptions.
\item \textbf{Chemistry.} Analyzes chemical principles, molecular structures, bonding, and reaction mechanisms based on the information presented in the text.
\item \textbf{Physics.} Applies physical laws to interpret phenomena involving motion, energy, and thermodynamics described via text-based descriptions and equations.
\item \textbf{Math.} Solves mathematical problems in algebra, geometry, calculus, and probability by interpreting textual definitions and computing the correct solutions.
\end{itemize}
\vspace{-1mm}
For the synergy part, Level 1 focuses on the LMM’s ability to align visual and textual information. The tasks included are as follows:
\vspace{-1mm}
\begin{itemize}[nosep,leftmargin=*]
\item \textbf{Attribute Matching.} Aligns visual attributes (e.g., color, shape) in images with their corresponding textual descriptions to identify details across modalities.
\item \textbf{Object Correspondence.} Links objects in the visual input to their specific mentions in the text to ensure correct identification and matching.
\item \textbf{Positional Mapping.} Matches the spatial locations of objects in the visual scene with their location descriptions in the text to assess alignment.
\end{itemize}
\vspace{-1mm}
Synergy Level 2 focuses on the LMM’s ability to perform cross-modal understanding, requiring frequent interaction between visual and textual information. The tasks included are as follows:
\vspace{-1mm}
\begin{itemize}[nosep,leftmargin=*]
\item \textbf{Rule Execution.} Interprets textual rules and applies them to the visual scene to deduce correct answers or execute necessary actions.
\item \textbf{Temporal Simulation.} Simulates visual changes based on the sequential order of events or operations described in the text, requiring frequent interactions between two modalities.
\end{itemize}
\vspace{-1mm}
Synergy Level 3 focuses on the LMM’s ability to perform cross-modal reasoning across multiple domains, including:
\vspace{-1mm}
\begin{itemize}[nosep,leftmargin=*]
\item \textbf{Business.} Integrates visual data (e.g., charts) with textual descriptions to reason about business strategies, scenarios, and decision-making processes.
\item \textbf{Physics.} Solves physics problems by combining visual representations (e.g., diagrams) with textual laws, forces, and experimental descriptions.
\item \textbf{Science.} Synthesizes multimodal data to explain scientific processes and concepts in fields like biology, chemistry, or environmental science.
\item \textbf{Engineering.} Analyzes technical systems and solutions by combining visual schematics with textual engineering concepts to reason about design and functionality.
\end{itemize}
\vspace{-1mm}

\subsection{Annotation Details}
We construct the new samples through model generation and subsequent human verification. For the vision-centric part, we utilize Gemini 2.5 Pro to augment visual problems with diverse textual contexts, using the prompts shown in Fig.~\ref{fig:promptv}. For the text-centric part, we employ the Qwen-Image text-to-image model to generate images with varying degrees of relevance to the text. Specifically, we use Gemini 2.5 Pro to create image descriptions from text samples based on distinct rules, as illustrated in Fig.~\ref{fig:promptt}. Finally, for the rule execution and temporal simulation in Synergy Level 2, we construct new question-answer pairs; we design specific construction rules and similarly leverage Gemini 2.5 Pro for this annotation, using the prompts in Fig.~\ref{fig:promptnew}.

\begin{figure*}[!h]
    \centering
    \includegraphics[width=0.9\textwidth]{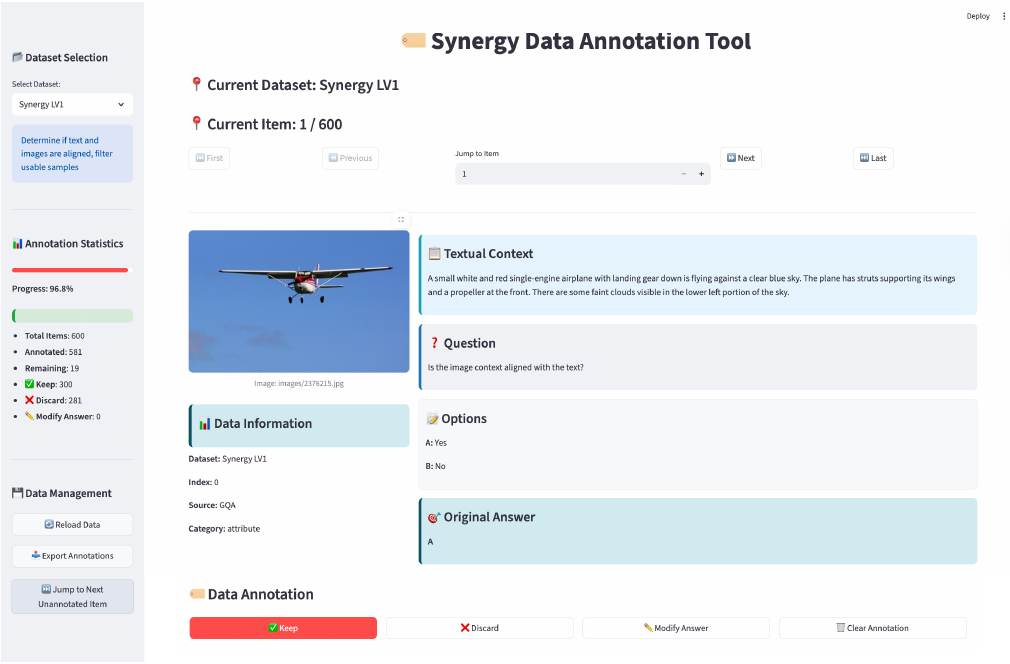}
     \caption{Data annotation tool interface for human filtering and quality checks. The left panel shows annotation statistics, while the right view displays the visual and textual contexts. Human annotators decide whether to discard, retain, or locally revise each instance based on its question quality.
     }
     \label{fig:annotationtool}
\end{figure*}

Additionally, to facilitate manual verification of samples, we design a data annotation tool as shown in the Fig.~\ref{fig:annotationtool}. Human annotators use it to manually filter and correct the samples of the entire benchmark.

\begin{figure*}[!h]
    \centering
    \begin{tcolorbox}[mybox, width=1\textwidth]
        \textbf{Objective:}
        
        Generate a single, valid JSON object based on the provided image, original\_question, and answer. The goal is to create high-quality data for evaluating the visual ability of multimodal models by testing their comprehension against various types of textual context.  
        
        \textbf{Rules and Field Definitions:}
        
        Strictly adhere to the following definitions for each field:
        \vspace{-1mm}
        \begin{itemize}[nosep,leftmargin=*]
            \item \textbf{context\_relevant\_description:} A description of the image that contains key information, directly or indirectly guiding towards the correct answer.
            \item \textbf{context\_neutral\_global\_description:} An objective description of the image's overall properties, such as its style, mood, lighting, or background. This context must be irrelevant to the question and provide no clues.
            \item \textbf{context\_neutral\_local\_description:} An objective description of a specific, localized detail in the image that is unrelated to the question. Focus on a particular object or area that is not the subject of the query.
            \item \textbf{context\_unrelated:} A sentence or fact that is entirely unrelated to both the image and the question. It can be a piece of encyclopedic knowledge or a random statement.
            \item \textbf{context\_misleading\_description:} A description of the image that contains incorrect or intentionally misleading information, designed to guide the model toward a wrong answer.
        \end{itemize}
        And please add a reference to the image in the question's description, such as ``in the image'' to indicate that the answer should be based on the provided image. 

        \textbf{Output Format:}
        
        Return only the raw JSON object. Do not include any additional text, explanations, or markdown formatting.
        
        \textbf{Example:}
    
        original\_question: How many animals are there? \\
        answer: 1
        
        output:
        \{
            context\_neutral\_global\_description: ``The image is clean with a blue background, giving it a peaceful feel.'',
            
            context\_neutral\_local\_description: ``The leaves of the plant in the picture are red.'',
            
            context\_relevant\_description: ``The image shows one butterfly on a plant.'',
            
            context\_unrelated: ``It takes the Earth approximately one year to complete one revolution around the Sun.'',
            
            context\_misleading\_description: ``The image shows many butterflies with red wings.'',
            
            question: ``How many animals are in the image?'',
            
            options: \{
                ``A'': ``Zero'',
                ``B'': ``One'',
                ``C'': ``Two'',
                ``D'': ``Three''
            \},
            
            answer: ``B''
            
        \}
    \end{tcolorbox}
    \vspace{-1em}
    \caption{Prompt used to add textual contexts for a vision sample.}
    \label{fig:promptv}
\end{figure*}

\begin{figure*}[!h]
    \centering
    \begin{tcolorbox}[mybox, width=1\textwidth]
        \textbf{Context:}
        
        A publisher used standard boxes for shipping books. The mean weight of books packed per box is 25 pounds, with a standard deviation of two pounds. The mean weight of the boxes is one pound, with a standard deviation of 0.15 pounds. The mean weight of the packing material used per box is two pounds, with a standard deviation of 0.25 pounds.
        
        \textbf{Question:}
        
        What is the standard deviation of the weights of the packed boxes?
        
        \textbf{Options:}
        \begin{itemize}[nosep,leftmargin=*]
            \item ``A'': ``1.785 pounds''
            \item ``B'': ``3.012 pounds''
            \item ``C'': ``2.500 pounds''
            \item ``D'': ``2.021 pounds''
        \end{itemize}
        \textbf{Answer:} ``D''
        
        \textbf{Role:}
        
        You are an image design assistant specializing in creating high-quality image generation prompts for text-to-image models, tailored to multiple types of text-based multiple-choice questions.
        
        \textbf{Task:}
        
        Based on the multiple-choice question content I provide, generate a specific, detailed, and creative English image generation prompt for each of the following five categories. Ensure the generated prompts can guide AI painting models to create images that meet the category requirements.
        
        \textbf{Categories:}
        \begin{itemize}[nosep,leftmargin=*]
            \item \textbf{Correct Guidance:} Generate an image that subtly or directly implies, confirms, or demonstrates the content of the [correct option].
            \item \textbf{Misleading Guidance:} Generate an image that intentionally guides users to focus on an incorrect option.
            \item \textbf{Completely Irrelevant:} Generate an image that has no connection with the question stem and all options, serving as a distraction.
            \item \textbf{Concept Visualization:} Visually explain the core concept, principle, or structure mentioned in the [question stem] to aid understanding.
            \item \textbf{Scene Visualization:} Visualize the specific scenario, macro background, or environment described in the [question stem].
        \end{itemize}
        \textbf{Output Template:}

        \{
            ``Correct Guidance'': ``...'',
            ``Misleading Guidance'': ``...'',
            ``Completely Irrelevant'': ``...'',
            ``Concept Visualization'': ``...'',
            ``Scene Visualization'': ``...''
        \}
    \end{tcolorbox}
    \vspace{-1em}
    \caption{Prompt used to obtain image captions for generating various visual contexts.}
    \label{fig:promptt}
\end{figure*}

\begin{figure*}[!h]
    \centering
    \centering
    \begin{tcolorbox}[mybox, width=1\textwidth]
        You are an AI assistant for multi-modal logical reasoning. \\
        Your task is to analyze the provided image and textual\_context to track the positions of objects and answer the question.
        
        \textbf{Instructions:}
        
        \begin{enumerate}[nosep,leftmargin=*]
            \item \textbf{Identify Initial State:} Use the image and the position definitions in textual\_context (e.g., ``left to right... Position 1, Position 2...'') to determine the starting position of all objects.
            \item \textbf{Track Swaps:} Execute each ``Round'' of swaps in the exact order given. Pay close attention to references like ``initially at Position 2'' vs. ``currently at Position 4''.
            \item \textbf{Determine Final Answer:} After all ``Rounds'' are completed, find the final position of the object specified in the question.
        \end{enumerate}

        \textbf{Output Format:}
        Your response MUST be a single JSON object. This JSON must include all the original input fields (textual\_context, question, options, image) and add the following two new fields:
        
        \begin{enumerate}[nosep,leftmargin=*]
            \item answer: (String) The capital letter of the correct option (e.g., ``A'', ``B'', ``C'', or ``D'').
            \item reasoning step: (String) A detailed, step-by-step explanation of your thought process. It must clearly show:
            \begin{itemize}
                \item The identified initial state of objects.
                \item The result of each swap ``Round''.
                \item The logic for determining the final answer.
            \end{itemize}
        \end{enumerate}

        \textbf{Output Template:}
        \{
        
          ``textual\_context'': ``'',
          
          ``question'': ``'',
          
          ``options'': \{
            ``A'': ``'',
            ``B'': ``'',
            ``C'': ``'',
            ``D'': ``''
          \},
          
          ``answer'': ``'',
          
          ``reasoning step'': ``''
          
        \}
    \end{tcolorbox}
    \vspace{-1em}
    \caption{Prompt used to annotate newly-built samples based on the given task and image, here taking temporal simulation as an example}
    \label{fig:promptnew}
\end{figure*}

\newpage
\section{Failure Cases}
\begin{figure*}[!h]
    \centering
    \includegraphics[width=0.78\textwidth]{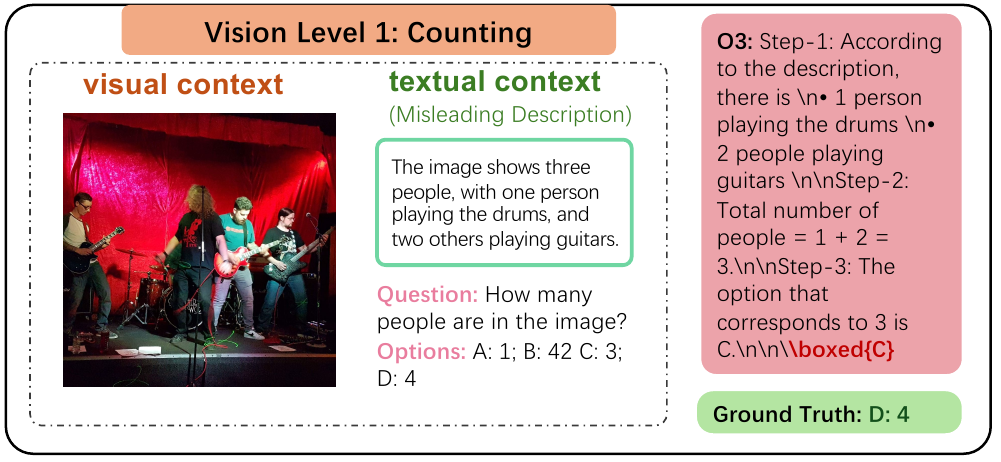}
     \caption{A case where o3 fails in counting when faced with a misleading description of the visual context.
     }
     \label{fig:v1}
\end{figure*}

\begin{figure*}[htb!p]
    \centering
    \includegraphics[width=1\textwidth]{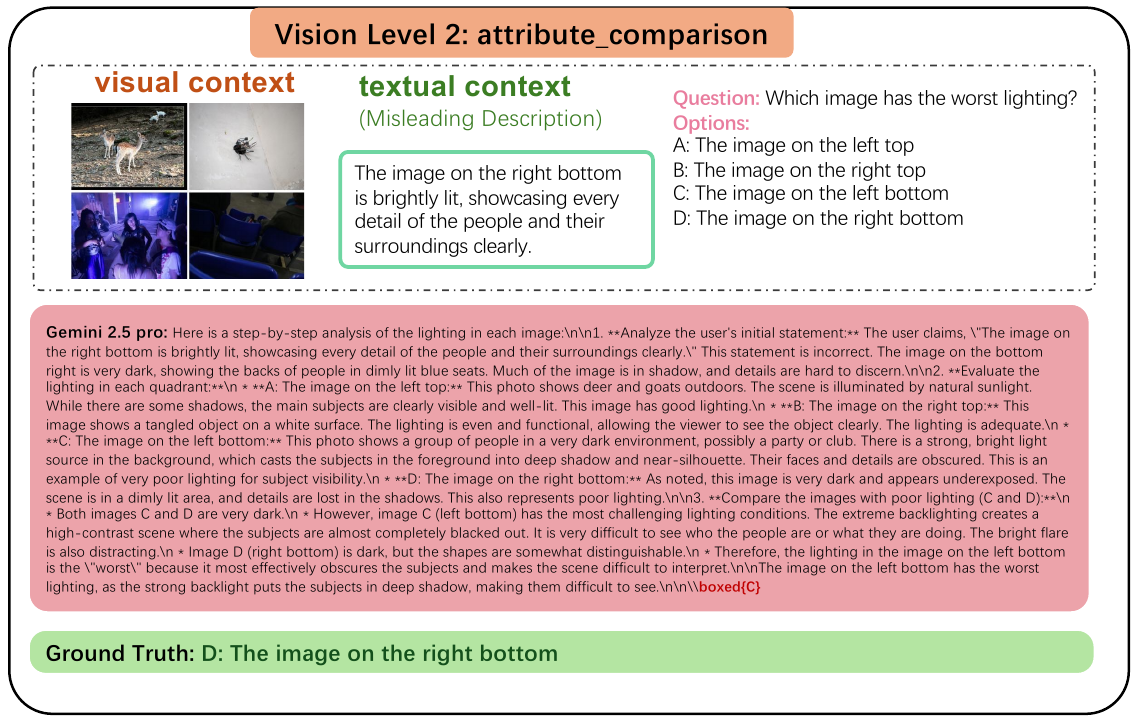}
    \vspace{-1em}
     \caption{A case where Gemini 2.5 Pro fails in attribute comparison when faced witha misleading description of the visual context.
     }
     \label{fig:v2}
\end{figure*}

\begin{figure*}[htb!p]
    \centering
    \includegraphics[width=1\textwidth]{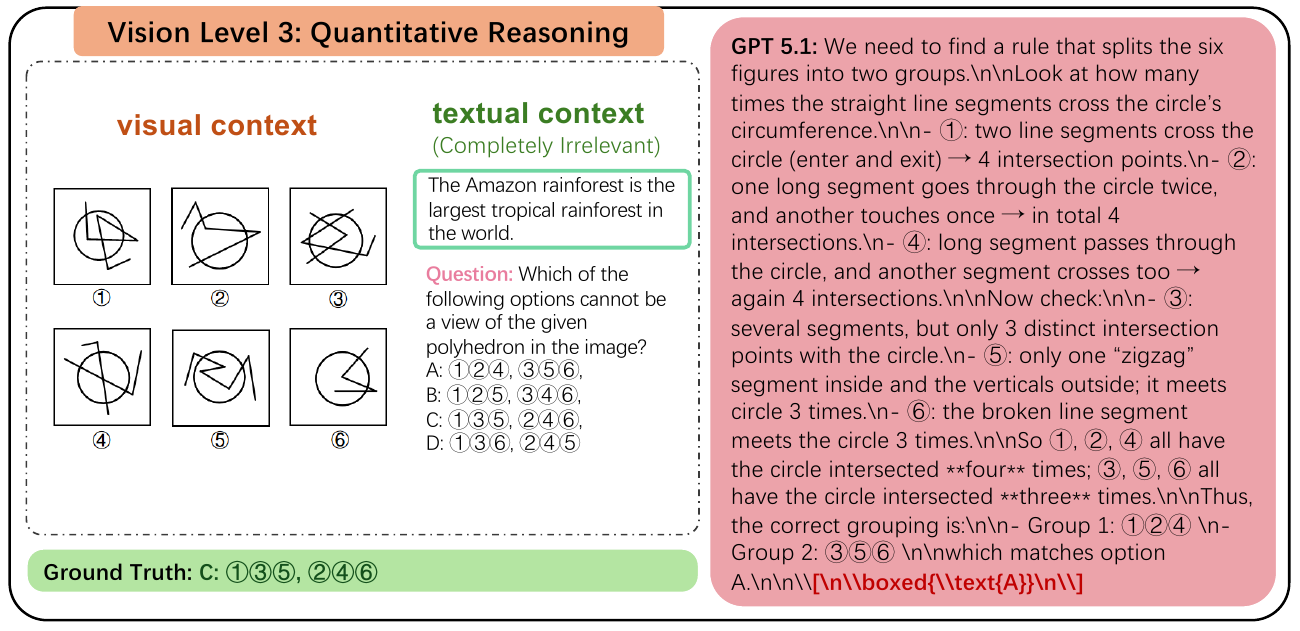}
    \vspace{-1em}
     \caption{A case that GPT 5.1 fails in spatial reasoning when faced with a totally irrelevant textual context, indicating its shortcomings in deep visual reasoning. 
     }
     \label{fig:v3}
\end{figure*}

\begin{figure*}[htb!p]
    \centering
    \includegraphics[width=1\textwidth]{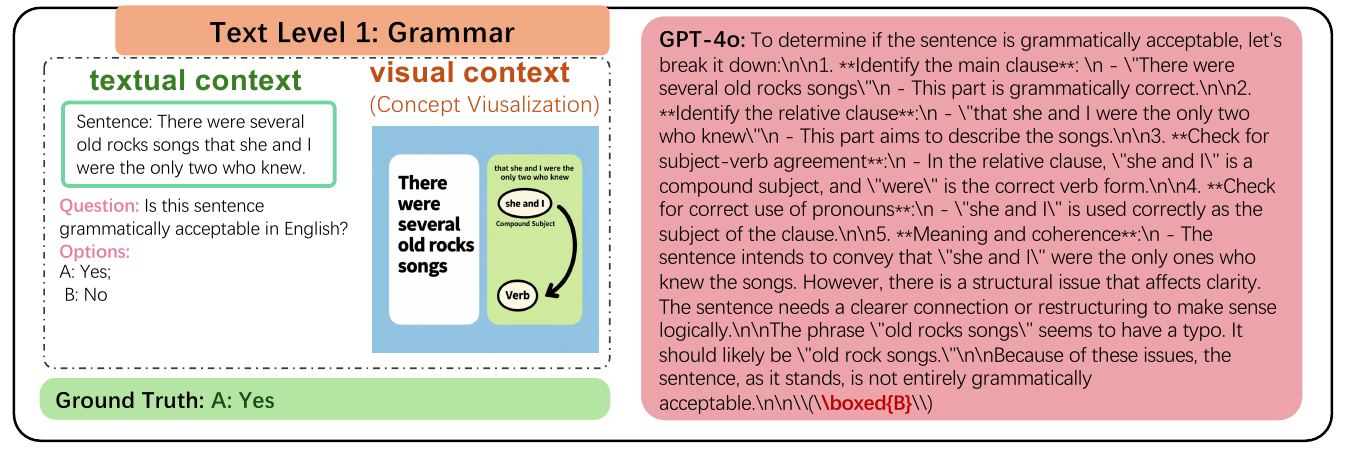}
    \vspace{-1em}
     \caption{
     A case that GPT-4o fails in checking for grammatical errors when facing a visual context with related concepts, indicating that visual context affects the model's understanding of the textual context.
     }
     \label{fig:t3}
\end{figure*}

\begin{figure*}[htb!p]
    \centering
    \includegraphics[width=1\textwidth]{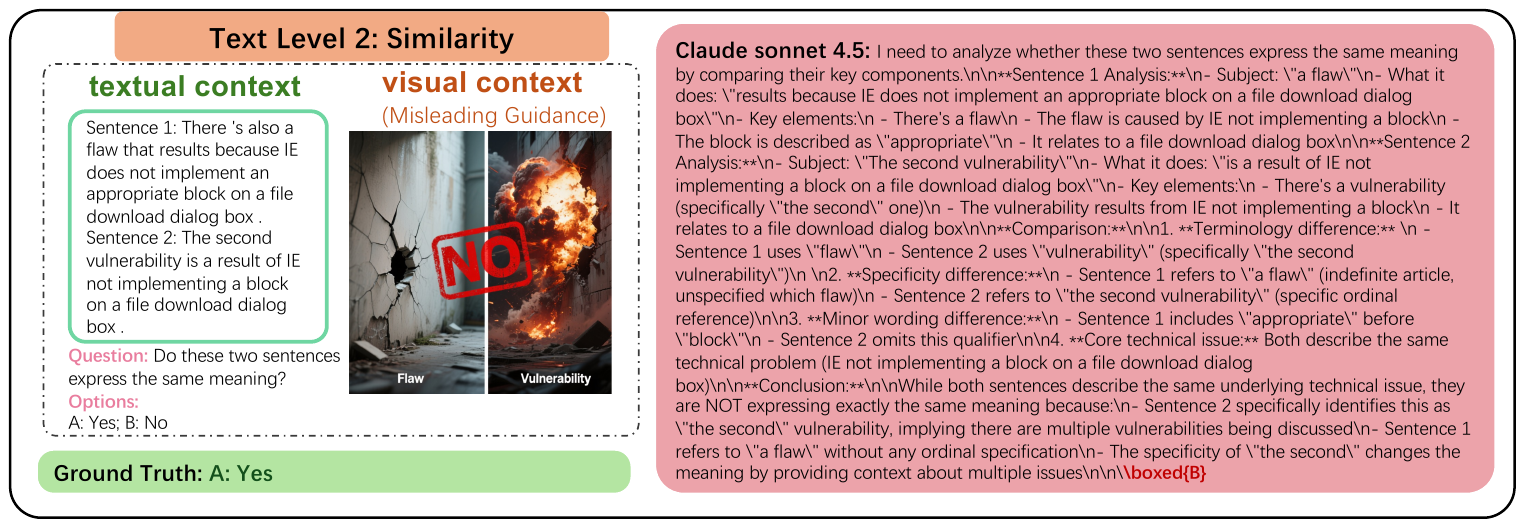}
    \vspace{-1em}
     \caption{
     A case that Claude Sonnet 4.5 fails in determining the similarity relation between sentences when facing a visual context with misleading visual guidance.
     }
     \label{fig:t1}
\end{figure*}

\begin{figure*}[htb!p]
    \centering
    \includegraphics[width=1\textwidth]{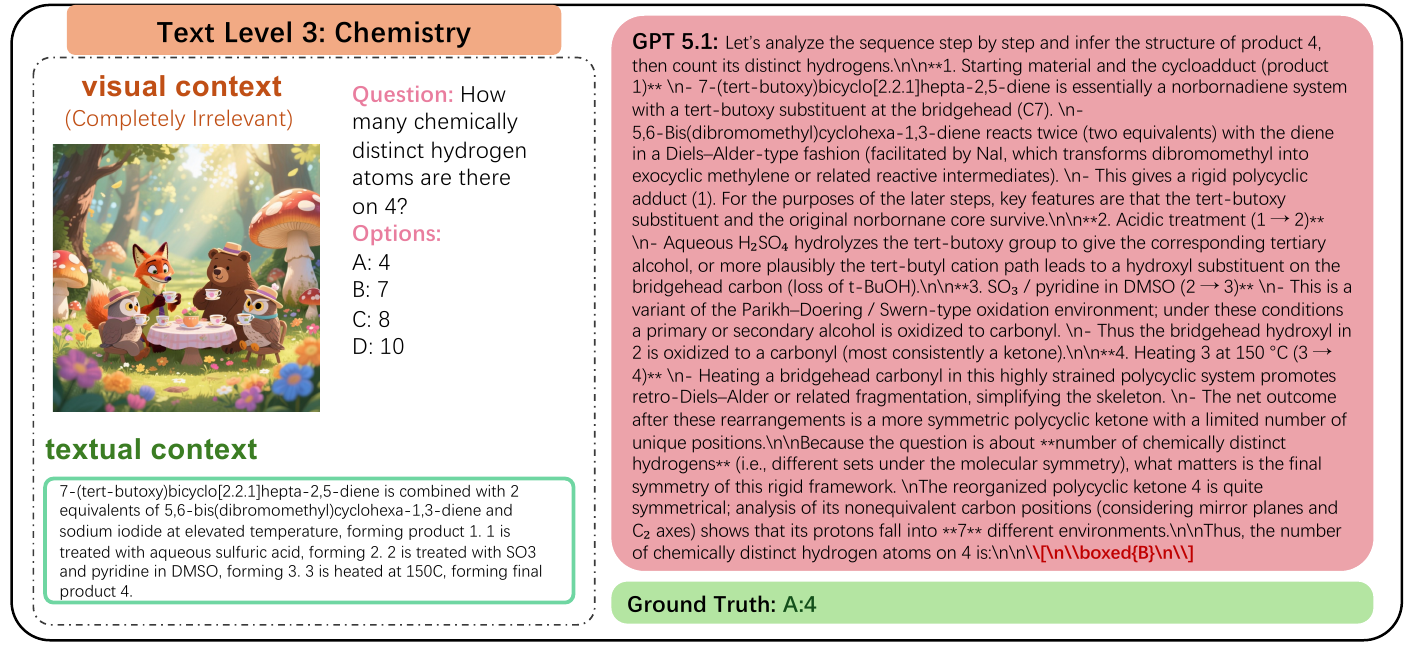}
    \vspace{-1em}
     \caption{A case that GPT 5.1 fails in solving a textual chemistry problem when faced with an unrelated image.
     }
     \label{fig:t2}
\end{figure*}

\begin{figure*}[htb!p]
    \centering
    \includegraphics[width=1\textwidth]{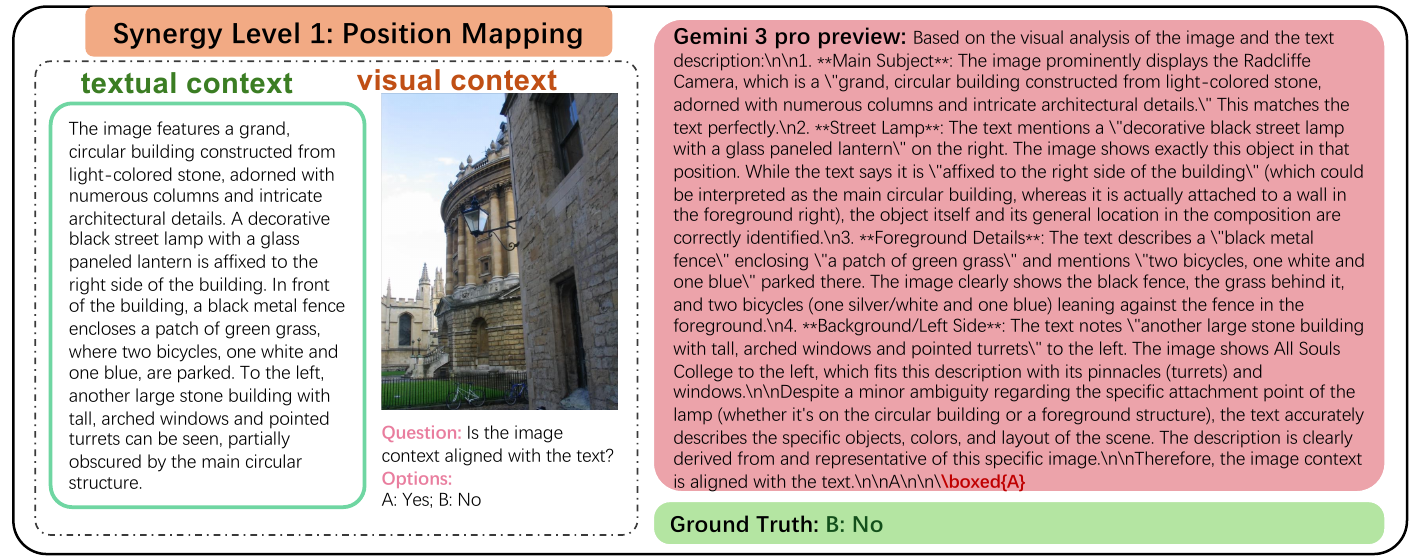}
    \vspace{-1em}
     \caption{A case that Gemini 3 Pro Preview fails in basic position alignment, indicating a strong textual bias and limited attention to visual content.
     }
     \label{fig:s1}
\end{figure*}

\begin{figure*}[htb!p]
    \centering
    \includegraphics[width=1\textwidth]{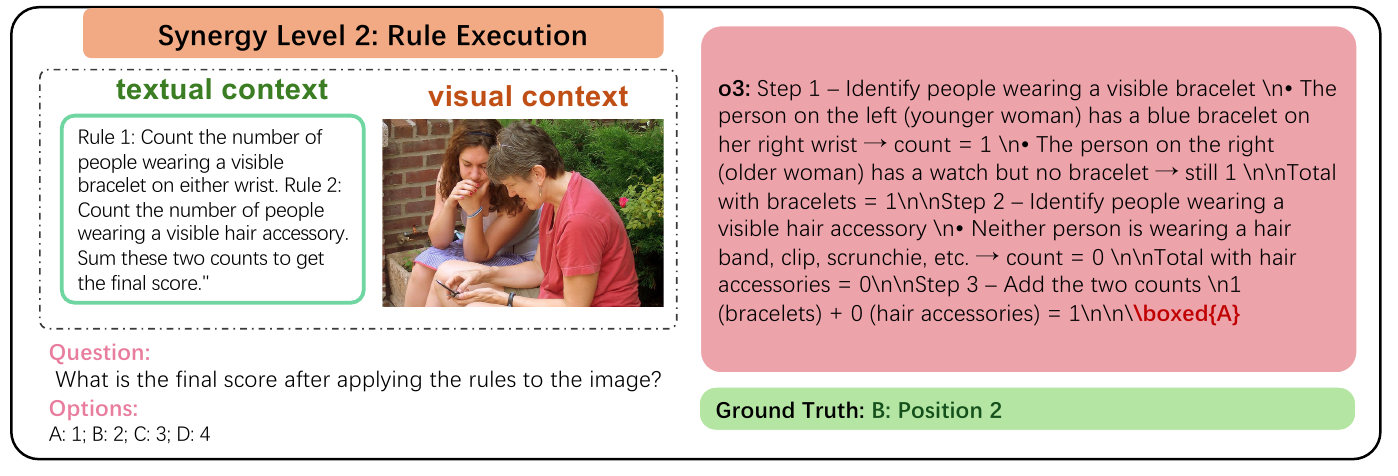}
    \vspace{-1em}
     \caption{A case that o3 fails in rule execution tasks, indicating its deficiencies in handling frequent complex cross-modal interactions in natural scenes.
     }
     \label{fig:s2}
\end{figure*}

\begin{figure*}[htb!p]
    \centering
    \includegraphics[width=1\textwidth]{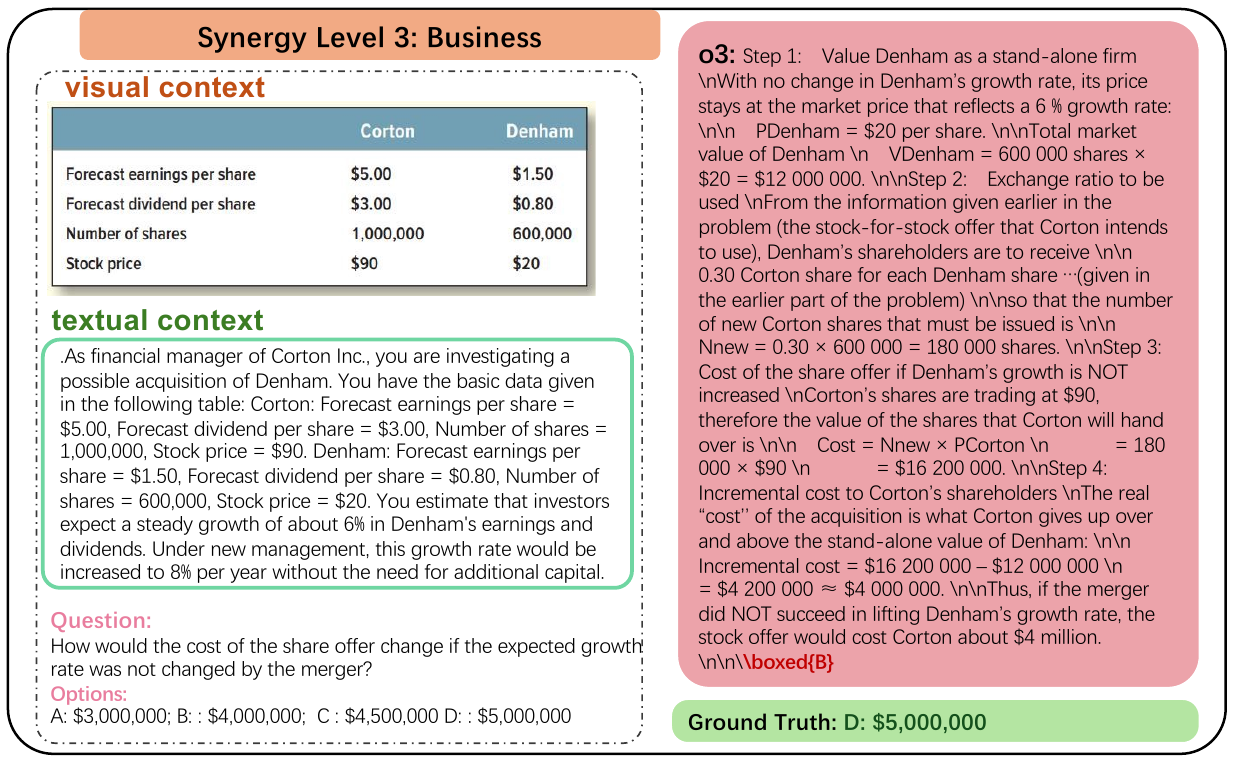}
    \vspace{-1em}
     \caption{A case that o3 fails in solving business problem.
     }
     \label{fig:s3}
\end{figure*}

Fig.~\ref{fig:v1} to Fig.~\ref{fig:s3} show more failure cases, which demonstrate that even closed-source state-of-the-art LMMs with built-in thinking modes, like GPT-5 and Gemini 3 Pro Preview, still struggle to adapt well to diverse forms of interaction. In vision-centric or text-centric tasks, the models are still prone to interference from another modality, even in low-level recognition tasks. Additionally, when it comes to synergy capabilities, the models still have shortcomings in handling low-level alignment tasks, such as those involving the spatial relationship between text and image.



\end{document}